\documentclass[journal]{IEEEtran}

\usepackage{siunitx}
\usepackage{amssymb}
\usepackage{bm}
\usepackage{amsmath}
\usepackage{soul}
\usepackage{rotating}
\usepackage{lscape}
\usepackage[export]{adjustbox}
\usepackage{graphics}

\usepackage{array}
\usepackage{floatrow}
\usepackage{multirow}
\usepackage{subcaption}
\usepackage{tabu}
\usepackage{algorithm}
\usepackage{algpseudocode}

\usepackage{todonotes}

\usepackage{marginnote}

\newcommand{\xb}{\textbf{x}} 
\newcommand{\yb}{\textbf{y}} 
\newcommand{\zb}{\textbf{z}} 
\newcommand{\gb}{\textbf{g}} 
\newcommand{\imb}{\textbf{m}} 
\newcommand{\mb}{\bm{\mu}} 
\newcommand{\Sib}{\bm{\Sigma}}
\DeclareMathOperator{\E}{\mathbb{E}}
\DeclareMathOperator*{\argmin}{arg\,min}
\DeclareMathOperator*{\argmax}{arg\,max}

\newcommand{\beginsupplement}{%
        \setcounter{table}{0}
        \renewcommand{\thetable}{S\arabic{table}}%
        \setcounter{figure}{0}
        \renewcommand{\thefigure}{S\arabic{figure}}%
     }
     
\newcommand{\reva}[1]{\textcolor{black}{#1}}
\newcommand{\revb}[1]{\textcolor{black}{#1}}
\newcommand{\revc}[1]{\textcolor{black}{#1}}

\begin{document}

\title{MR image reconstruction using deep density priors}


\author{\IEEEauthorblockN{Kerem C. Tezcan,
Christian F. Baumgartner, Roger Luechinger, Klaas P. Pruessmann,
Ender Konukoglu}
\thanks{Manuscript received March 19, 2018; revised November 27, 2018. The presented work is funded by Swiss National Science Foundation grant number: 205321\_173016. KCT, CFB and EK are with the Computer Vision Laboratory, ETH Z\"urich, Switzerland. Emails: \{tezcan, baumgartner, ender.konukoglu\}@vision.ee.ethz.ch. RL and KPP are with the Institute for Biomedical Engineering, ETH Z\"urich and University of Z\"urich, Switzerland. Emails: \{luechinger, pruessmann\}@biomed.ee.ethz.ch. Copyright (c) 2018 IEEE. Personal use of this material is permitted. However, permission to use this material for any other purposes must be obtained from the IEEE by sending a request to pubs-permissions@ieee.org. Supplementary materials are available in the supplementary files / multimedia tab. doi:10.1109/TMI.2018.2887072. https://ieeexplore.ieee.org/document/8579232}}

\markboth{IEEE TMI, 2018, \MakeLowercase{doi}:10.1109/TMI.2018.2887072 }%
{Shell \MakeLowercase{\textit{et al.}}: Bare Demo of IEEEtran.cls for IEEE Transactions on Magnetics Journals}
%



\IEEEtitleabstractindextext{%
\begin{abstract}
	Algorithms for Magnetic Resonance (MR) image reconstruction from undersampled measurements exploit prior information to compensate for missing k-space data. 
	Deep learning (DL) provides a powerful framework for extracting such information from existing image datasets, through learning, and then using it for reconstruction. 
	\reva{Leveraging this, recent methods employed DL to learn mappings from undersampled to fully sampled images using paired datasets, including undersampled and corresponding fully sampled images, integrating prior knowledge implicitly. 
	%
	In this article, we propose an alternative approach that learns the probability distribution of fully sampled MR images using unsupervised DL, specifically Variational Autoencoders (VAE), and use this as an explicit prior term in reconstruction, completely decoupling the encoding operation from the prior. 
	The resulting reconstruction algorithm enjoys a powerful image prior to compensate for missing k-space data without requiring paired datasets for training nor being prone to associated sensitivities, such as deviations in undersampling patterns used in training and test time or coil settings.}
	We evaluated the proposed method with T1 weighted images from a publicly available dataset, multi-coil complex images acquired from healthy volunteers (N=8) and images with white matter lesions. 
	The proposed algorithm, using the VAE prior, produced visually high quality reconstructions and achieved low RMSE values, outperforming most of the alternative methods on the same dataset. 
	On multi-coil complex data, the algorithm yielded accurate magnitude and phase reconstruction results. 
	In the experiments on images with white matter lesions, the method faithfully reconstructed the lesions.
\end{abstract}

\begin{IEEEkeywords}
Reconstruction, MRI, prior probability, machine learning, deep learning, unsupervised learning, density estimation
\end{IEEEkeywords}}

\maketitle

\IEEEdisplaynontitleabstractindextext

\section{Introduction}

	\IEEEPARstart{A}{cquisition} time in magnetic resonance (MR) imaging is directly related to the number of samples acquired in k-space. For high quality images, a large number of samples, and therefore long acquisition times are necessary. Reducing acquisition time in a reliable manner is an important question in MR imaging and many methods exploiting different properties of the k-space~\cite{partialfourier} and hardware design~\cite{sense} have found wide use in clinical practice. 

	The question also received considerable attention from the image and signal processing research communities, main focus being reconstruction methods from randomly or regularly undersampled k-space acquisitions. The random undersampling approach was primarily motivated by the compressed sensing framework~\cite{Lustig2007,Guerquin-kern11afast, Lingala2011,donoho, candesetal}, where the incoherence between k-space sampling and some sparsifying transform was exploited. 
	Some works used hand-crafted sparsifying transformations, such as gradient in total variation~\cite{RUDIN1992259} (TV) and wavelet decomposition~\cite{ning_sidwt, qu_pbdw}, while others used dictionary learning to determine the transformation from data~\cite{dlrecon,zhan_fdlcp}. 
	An alternative approach in~\cite{Eksioglu2016} used block matching and 3D filtering, based on the BM3D denoising method~\cite{Dabov_bm3d}, to tackle reconstruction as a denoising problem and exploit similarity of image patches to determine a sparsifying transform (BM3D-MRI). 
	On the other hand, regular undersampling schemes and corresponding reconstruction algorithms were also extensively investigated~\cite{Wissmann2017,KlaasP.Pruessmann2001, Pedersen2009, Schmidt2014}. 
	The common aspect in all of these approaches, notably except BM3D-MRI, is that they invert underdetermined and ill-posed systems of equations by using good regularizers, which can be viewed as introducing explicit prior information on the expected structure that helps eliminate artifacts.

	Recently, researchers started to employ deep neural networks (DNN)~\cite{deeplearning} for MR reconstruction~\cite{litjenssurvey}. 
	Two main approaches have been explored. The first is to learn a mapping from undersampled images to the fully sampled versions. 
	For a given undersampling pattern,~\cite{automap, kwon_mlp}~\revb{\cite{lee}} proposed to learn a feed-forward network and during test time fed the undersampled image to the network. 
	Similarly, authors in \cite{2017arXiv170301135H} proposed to train a  network to remove spiking artifacts in computed tomography images and modified the mapping for MR using domain adaptation.  
	The feed-forward network strategy, however, could not guarantee data consistency at test time. 
	To address this, authors in~\cite{cascnet} added explicit data-consistency. 
	Here, a mapping block, based on Convolutional Neural Networks (CNN), is followed by a data consistency block and the dual block structure is cascaded. 
	This strategy guaranteed consistency with measured data during test time while the CNNs perform de-aliasing. 
	Alternatively, authors in \cite{wang} proposed a model where a feed-forward mapping is learned and during test time its output is used as a regularization term alongside the data consistency and other regularizers. 
	The aim was to make sure the final reconstruction did not deviate too much from the mapping output.

	The second DL approach leverages networks to improve existing iterative reconstruction algorithms in terms of reconstruction quality. 
	In \cite{NIPS2016_6406} the authors showed that the iterations of the alternating direction method of multipliers (ADMM) algorithm can be unrolled as a multi-layer CNN.
	Instead of being fixed as in the original method, kernels and non-linear functions are parameterized and learned by the network, which improved reconstruction accuracy. 
	A similar method was presented in \cite{hammernikvar} applying the technology developed in \cite{chendiff}, which used the same principle as~\cite{NIPS2016_6406} for improving diffusion filtering. 
	\revb{The same strategy of expressing an existing iterative method as a convolutional network was also used in~\cite{unser_dcnn, jacob} and in~\cite{putzky} using recurrent neural networks.} 
	\reva{Algorithms taking either of the DL approaches also integrate prior information on the structure of fully sampled images to account for missing k-space measurements but in an implicit manner, embedded in the trained weights of the feed-forward networks.}

	In this work, we employ a neural network as an explicit prior similar in essence to the non-DL based reconstruction methods proposed in the signal processing communities, the difference being the power of the prior model. Unsupervised learning with DNNs has been very successful in approximating probability distributions of high dimensional data, including images, from a set of samples. One such approach, of particular interest to the method proposed here, is the variational auto encoder (VAE) algorithm~\cite{KingmaW13, rezendevae}. Using VAEs, it is possible to approximate the distribution of MR image patches and likelihood of a previously unseen image. Furthermore, the approximate likelihood function is a network and therefore differentiable. These two aspects allow using a VAE as a prior model that can approximate distributions of large image patches, e.g. patches of 28x28 pixels, for iterative reconstruction.

	We propose a novel probabilistic reconstruction method that uses priors learned via VAEs, which we term as Deep Density Prior (DDP) based reconstruction. We formulate a Bayesian model of the imaging process, including the prior and a data consistency term that embeds the encoding operation, and express DDP as the Maximum-A-Posteriori (MAP) estimation. \reva{Compared to non-DL based methods, the main difference of DDP is the powerful prior that can capture distribution of large image patches. Compared to DL approaches, the fundamental differences of DDP are: (i) the explicit prior, which is trained to capture the structure of the fully sampled images as opposed to the implicit prior in feed-forward networks that is learned to reduce artifacts seen during training; and (ii) the decoupling of the prior from the data consistency term.} The latter difference leads to two theoretical advantages: \reva{(i) while previously proposed DL-based approaches required paired datasets for training, including undersampled and corresponding fully sampled images, DPP does not}; and \revb{(ii) decoupling prior and data consistency terms eliminates possible sensitivities in accuracy to deviations in acquisition specifications between undersampled images used for training and encountered during test time, such as sampling parameters, coil settings and k-space trajectories,} \reva{which for instance has been demonstrated to be the case in~\cite{automap}.}

	In the rest of article, we first present the method and then show reconstruction results and comparisons with conventional approaches as well as recent DNN based methods. 
\section{Methods}
	In the first two parts of this section, we provide a brief background on Bayesian formulation of the MR reconstruction problem and the VAE algorithm. We present our main technical contribution, learning a prior for MR patches and integrating it in the reconstruction problem, starting from Section~\ref{sec:derivative}.

	\subsection{Bayesian formulation of the MR reconstruction problem}
	An MR image is denoted as $\imb \in \mathbb{C}^N$, where N is the number of pixels\footnote{In this work we focus on 2D imaging, however, the same techniques can be applied to 3D imaging and this extension will be a part of future work.}. An imaging operation is given by an undersampling encoding operation $E=UFS$, where $S : \mathbb{C}^N \rightarrow \mathbb{C}^{N\times\gamma}$ is a sensitivity encoding operator. $\gamma$ is the number of coils, $F: \mathbb{C}^{N\times\gamma} \rightarrow \mathbb{C}^{N\times\gamma}$ is the Fourier operator and $U: \mathbb{C}^{N\times\gamma} \rightarrow \mathbb{C}^{M\times\gamma}$ is an undersampling operator, with $M<N$. Let us also define $\xb\in\mathbb{C}^P$ as an image patch of $P$ pixels extracted from $\imb$. 

	Assuming complex-valued, zero mean, normal distributed and uncorrelated additive noise, denoted as $\eta$, the acquired data $\yb \in \mathbb{C}^{M\times\gamma}$ can be modelled as $\yb = E\imb + \eta$. Under this noise model the data likelihood becomes
	\begin{equation}\label{eqn:data_likelihood}
	p(\yb|\imb) = \mathcal{N}(\yb|E\imb, \sigma_{\eta}) = \frac{1}{(2\pi\sigma_{\eta}^2)^{M/2} } e^{-\frac{1}{2\sigma_{\eta}^2}(E\imb-\yb)^H (E\imb-\yb)},
	\end{equation}
	where H denotes the Hermitian transpose and $\sigma_{\eta}$ is the standard deviation of the noise. In reality, the noise might have some correlation especially in multi-coil acquisitions, which can be corrected by prewhitening. Here, in order to keep things simple, we neglect the noise correlation. In reconstruction, the quantity of interest is the posterior distribution $p(\imb | \yb)$, i.e. the probability of the image being $\imb$ given the k-space measurements. A common approach to model the reconstruction problem is to use the MAP estimation
	\begin{equation}
	\argmax_\imb p(\imb|\yb) = \argmax_\imb \left[p(\yb|\imb) p(\imb)\right],
	\end{equation}
	where we used the Bayes' theorem and dropped the term $p(\yb)$ since it does not depend on $\imb$. $p(\imb)$ is called the prior term and represents the information one has about the fully sampled image before the data acquisition. Taking the $\log$ of both sides yields
	\begin{multline}\label{eqn:bayesrecon}
	\argmax_\imb \log p(\imb | \yb) = \argmax_\imb \left[\log p(\yb|\imb) + \log p(\imb)\right] \\ = \argmax_\imb \left[- \frac{1}{2\sigma_{\eta}} \lVert E\imb-\yb \rVert_2^2  + \log p(\imb)\right].
	\end{multline}
	Taking the maximum (or equivalently taking the minimum of the negative of the expression), defining the constant $\lambda \triangleq 2\sigma_{\eta}$ and multiplying both terms with it recovers the conventional formulation of a reconstruction problem with a data consistency and a regularization term that is weighted by the trade-off parameter~$\lambda$
	\begin{equation}\label{eqn:conventional}
	\hat{\imb} = \argmin_{\imb} \left[\lVert E\imb-\yb \rVert_2^2 - \lambda \log p(\imb)\right].
	\end{equation}

	In this work, we propose to estimate the prior term from examples of fully sampled images and approximate $-\log p(|\xb|)$, i.e. the negative log prior of magnitude of image patches, with a neural network model.  We train a VAE on patches extracted from fully sampled MR images to capture the distribution and use this prior for reconstruction. This allows us to utilize the prior independent of the sampling operation in contrast to the feed-forward mapping approach.

	\subsection{Learning the data distribution with VAEs} 
	VAE is an unsupervised learning algorithm proposed to approximate high-dimensional data distributions~\cite{KingmaW13, rezendevae}. We introduce VAEs very briefly\footnote{Brief explanation is due to space restrictions} and refer the reader to~\cite{KingmaW13} for further details. VAE is a generic algorithm that can be applied to any signal but we focus on the magnitude image patches in our description. 

	The main goal of the VAE algorithm is to approximate the data distribution using a latent variable model and optimize its parameters for a given set of examples using variational approximation. The model is given as
	\begin{equation}
	p(|\xb|) = \int_Z p(|\xb|,\zb) d\zb = \int_Z p(|\xb||\zb) p(\zb) d\zb,
	\end{equation}
	where $\zb \in \mathbb{R}^L$ denotes the latent variable, $p(\zb)$ the prior over the $\zb$'s and $L << P$. A known distribution is assumed for $p(\zb)$, e.g. unit Gaussian, and a parameterized $p(|\xb||\zb)$  is optimized to maximize $\log p(|\xb|)$ of observed samples. This modeling strategy is also taken in other probabilistic latent variable models, such as probabilistic principal component analysis~\cite{Bishopbook}. The VAE model parameterizes $p(|\xb||\zb)$ as a neural network whose set of parameters we denote with $\varphi$. To optimize $\log p(|\xb|)$ for the given samples, the integral over $\zb$ needs to be evaluated and this is not feasible even for moderate $L$. Variational approximation uses an approximate distribution for the posterior $q(\zb||\xb|) \approx p(\zb||\xb|)$ to address this problem. Using $q(\zb||\xb|)$, $\log p(|\xb|)$ can be decomposed into two terms~\cite{Bishopbook}
	\begin{multline}\label{eqn:log_likelihood}
	\log p(|\xb|)  = \E_{q(\zb||\xb|)}\left[\log\frac{p(|\xb|,\zb)}{q(\zb||\xb|)}  \right] + \text{D$_{\text{KL}}$} \left[ q(\zb||\xb|)  || p(\zb||\xb|) \right].
	\end{multline}
	The first term is referred to as the evidence lower bound (ELBO) and the second term is the Kullback-Leibler divergence (KLD) between the approximate and true posteriors. The KLD term \revb{is intractable because the true posterior $p(\zb||\xb|)$ is intractable. It is, however,} always larger than or equal to zero, which makes ELBO a lower bound for $\log p(|\xb|)$. The strategy of VAE is to maximize the ELBO as a proxy to $\log p(|\xb|)$.

	Similar to $p(|\xb||\zb)$, the VAE algorithm models $q(\zb||\xb|)$ as a separate neural network with parameters $\theta$ and during training optimizes both $\theta$ and $\varphi$ to maximize the ELBO of the training samples. Rewriting the ELBO with $p(|\xb||\zb)$, $p(\zb)$ and $q(\zb||\xb|)$, the optimization for training can be written as
	\small
	\begin{multline}\label{eqn:VAEoptimization}
	\max_{\theta, \varphi} \sum_{n=1}^N  \text{ELBO}(|\xb^n|)  \\ = \max_{\theta, \varphi} \left[\sum_{n=1}^{N} \E_{q_{\theta}(\zb||\xb^n|)}\left[\log p_{\varphi}(|\xb^n||\zb) \right] - \text{D$_{\text{KL}}$}\left[ q_{\theta}(\zb||\xb^n|) || p(\zb)\right]\right],
	\end{multline}
	\normalsize
	where $\xb^n$ is the $n^{th}$ training sample and we added the network parameters as subscript at the corresponding terms to indicate the dependence. Notice that the KLD term in Equation~\ref{eqn:VAEoptimization} is distinct from the one in Equation~\ref{eqn:log_likelihood}, it includes $p(\zb)$ instead of $p(\zb||\xb|)$, which makes it tractable.

	The networks $q_{\theta}(\zb||\xb|)$ and $p_{\varphi}(|\xb||\zb)$ are typically called the encoder and the decoder, respectively. The former takes a data sample $|\xb|$ and encodes it into a posterior distribution in the latent space with network parameters $\theta$. If the posterior distribution $q_{\theta}(\zb||\xb|)$ is modelled as a Gaussian, then the encoder outputs a mean and a covariance matrix for $\zb$ depending on $|\xb|$. The decoder network on the other hand, takes a latent vector $\zb$ and maps it to a conditional distribution of the data given $\zb$. During training, $\zb$ vectors are sampled from $q_{\theta}(\zb||\xb|)$ to evaluate the expectations in~\ref{eqn:VAEoptimization}. In this work, we use the original VAE design~\cite{KingmaW13} except for the data likelihood, for which we use a multi-modal Gaussian $p_{\varphi}(|\xb||\zb) = N(|\xb||\mb_{\varphi}(\zb),\Sib_{\varphi}(\zb))$ with a diagonal covariance matrix, similar to~\cite{epitome}. We note that the Gaussian distribution here is different from the data likelihood given in Equation~\ref{eqn:data_likelihood}. The Gaussian in Equation~\ref{eqn:data_likelihood} models the complex valued observation noise whereas the one here models conditional distribution of $|\xb|$ given $\zb$ in the essence of compound probability distributions. We provide further network design details in the supplementary materials.
	\subsection{Deep density prior (DDP) reconstruction model}\label{sec:derivative}
	Once the VAE model is trained we can integrate the prior within a Bayesian formulation of the reconstruction problem as given in Equation~\ref{eqn:bayesrecon}. We make two key observations to achieve this. First, given by the theory, the ELBO($|\xb|$) can be used as a proxy to the true distribution $\log p(|\xb|)$. So, an approximate log likelihood of a magnitude image patch $|\xb|$ can be obtained by evaluating $\text{ELBO}(|\xb|)$
	\begin{equation}
	\label{eqn:opt_elbo} \text{ELBO}(|\xb|) = \E_{q_{\theta^*}(\zb||\xb|)}\left[\log p_{\varphi^*}(|\xb||\zb) + \log \frac{p(\zb)}{q_{\theta^*}(\zb||\xb|)} \right],
	\end{equation}
	where $\theta^*$ and $\phi^*$ are the optimal VAE parameters learned during training. The approximate log-likelihood allows us to formulate the proposed reconstruction model as the following MAP estimation problem
	\begin{equation}\label{eqn:recon_opt_ideal}
	\argmin_{\imb} \left[\lVert E\imb-\yb \rVert_2^2 - \sum_{\xb_r\in{\Omega(\imb)}} \text{ELBO} \left( |\xb_r| \right)\right],
	\end{equation}
	where $\Omega(\imb)$ denotes a set of (overlapping) patches covering the image $\imb$ and $|\xb_r|$ is the magnitude of the $r^{th}$ image patch.  Note that this approach assumes independence between different patches, ignoring statistical dependencies between them. It would be possible to extend the model to achieve this, which is left for future work.

	Since an exact computation of the ELBO term requires evaluating the expectation with respect to $q(\zb||\xb|)$, which is computationally not feasible, we use a Monte Carlo sampling approach to calculate the ELBO as follows
	\begin{equation}
	\label{eqn:app_elbo} \text{ELBO}(|\xb|) \approx \frac{1}{J}\sum_{j=1}^{J} \log p(|\xb||\zb^j) + \log \frac{p(\zb^j)}{q(\zb^j||\xb|)}, \zb^j \sim q(\zb||\xb|).
	\end{equation}
	Here $J$ represents the number of Monte-Carlo samples.

	Plugging the ELBO approximation into Equation~\ref{eqn:recon_opt_ideal}, we obtain the formulation of the proposed DDP reconstruction problem
	\small
	\begin{equation}\label{eqn:recon_opt}
	\argmin_{\imb} 
	\lVert E\imb-\yb \rVert_2^2 - \sum_{\xb_r\in{\Omega(\imb)}} \Bigg[ \frac{1}{J}\sum_{j=1}^{J} \log p(|\xb_r||\zb^j) + \log \frac{p(\zb^j)}{q(\zb^j||\xb_r|)}\Bigg], 
	\end{equation}
	\normalsize
	where $\zb^j \sim q(\zb||\xb_r|)$, the first term is the usual data term and the second term within the summation is the regularization term that arises from the learned prior.

	Our second key observation is that the approximation in Equation~\ref{eqn:app_elbo} is differentiable since each term is defined through networks that are themselves differentiable. This is the critical aspect that allows integrating the trained VAE as a prior into an iterative reconstruction algorithm. We can compute the total derivative of the prior term with respect to each image patch as follows
	\begin{multline}
	\mathcal{R}(|\xb|, \zb^j) \triangleq \log p(|\xb||\zb^j) + \log \frac{p(\zb^j)}{q(\zb^j||\xb|)} \\
	\frac{d}{d\xb}\left[ \frac{1}{J}\sum_{j=1}^{J} \mathcal{R}(|\xb|, \zb^j)\right] = \frac{1}{J}\sum_{j=1}^{J}\frac{d}{d\xb}\mathcal{R}(|\xb|, \zb^j) \\
	= \frac{\xb}{|\xb|}\left[\frac{1}{J}\sum_{j=1}^{J}\frac{\partial}{\partial|\xb|}\mathcal{R}(|\xb|, \zb^j) + \frac{\partial}{\partial\zb^j}\mathcal{R}(|\xb|, \zb^j)\frac{d \zb^j}{d |\xb|}\right]\label{eqn:derivatives},
	\end{multline}
	where we defined $\mathcal{R}(|\xb|, \zb^j)$ for notational simplicity.  The second term in the last line is due to the dependency of the samples $\zb^j$ on $\xb$ and $\xb / |\xb|$ is due to taking the derivative of the magnitude with respect to the image patch. 
	\begin{algorithm}
	  \caption{Deep density prior (DDP) reconstruction using POCS. See text for a more detailed explanation.}
	  \begin{algorithmic}[1]
	    \item $y$: undersampled k-space data 
	    \item $E$: undersampling encoding operator
	    \item VAE: the trained VAE
	    \Procedure{DDPrecon}{$y$, $E$, VAE}
	      \State $\imb^0 \gets E^Hy$ \Comment{initialize with the zero-filled image}
	      \For{$t=0:T-1$}\Comment{main loop: POCS iterations}
	      \State $\imb^{t,0} \gets \imb^{t} $
	        \For{$k=0:K-1$}\Comment{inner loop: iterations for the prior projection $\mathcal{P}_{prior}$}
	            \State $\{\xb_r^{t,k}\} \gets \text{image2patches}(\imb^{t,k})$ \Comment{creates a set of patches covering the image}
	            \For{$r=1:$ no of patches} \Comment{loop over all the patches in $\{\xb_r^t\}$ }
	              \State $\gb_r \gets  \frac{d}{d\xb}\left[ \frac{1}{J}\sum_{j=1}^{J} \mathcal{R}(|\xb|, \zb^j)\right]_{\xb=\xb_r^{t,k}}$ \Comment{calculate the derivative acc. to Eq.~\ref{eqn:derivatives}}
	            \EndFor
	            \State $\gb \gets \text{patches2image}(\{\gb_r\})$ \Comment{$\mathcal{P}_{prior}\imb^t$ }
	            \State $ \imb^{t,k+1} \gets \imb^{t,k} + \alpha\cdot\gb  $ 
	            \State $\imb^{t,k+1} \gets  \mathcal{P}_{phase}\imb^{t,k+1}$ \Comment{(optional)}
	        \EndFor

	        \State $\imb^{t+1} \gets  \imb^{t,K} -  E^H (E\imb^{t,K} - y) $ \Comment{$\mathcal{P}_{DC}\imb^t$ }
	      \EndFor\label{euclidendwhile}
	      \State \textbf{return} $\imb^T$\Comment{Resulting reconstruction}
	    \EndProcedure
	  \end{algorithmic}
	  \label{alg:recon}
	\end{algorithm}
	\subsection{Optimization using projection onto convex sets}
	\label{sec:reconPOCS}

	We solve the DDP optimization problem given in Equation~\ref{eqn:recon_opt} using the projection onto convex sets (POCS) algorithm~\cite{SamsonovPOCS}, specifically using the formulation in~\cite{pocsunderrelaxed}. POCS is an iterative minimization process, where the solution variable is projected sequentially onto different convex sets, each defined by one of the constraints in the problem.

	\revb{The projection for the data consistency term is implemented using the method proposed in ~\cite{SamsonovPOCS}, which is $\mathcal{P}_{DC} \imb = \imb -  E^H (E\imb - \yb)$. When there are multiple coils, this projection implements SENSE reconstruction\footnote{For optimal signal-to-noise ratio, one needs to account for the noise covariance as well, which we ignore in this work for simplicity.}.} Since we do not have a projection operator for the prior term, we approximate it by several gradient ascent steps with a small step size $\alpha$ as in~\cite{pocsunderrelaxed}. We use the final image at the end of the ascent steps as the projected image patch. We define the prior projection with the following steps: i) create a set of patches $\{\xb_r^t\} = \Omega(\imb^t) $  from the image $\imb^t$ at iteration t, ii) obtain the derivatives for each of these patches using Equation~\ref{eqn:derivatives}, which have the same size as the patches themselves, iii) combine the derivatives of the patches to form a derivative image by averaging the values where the patches overlap, iv) update the image using the derivative image, v) repeat this K times. Notice that the set defined by the prior projection is not necessarily convex in $|\xb|$, however we have not encountered any problems in convergence during our experiments. To reduce edge effects resulting from patchwise projections, we use four sets of overlapping patches.

	With the data consistency and prior projections defined as above, one step of reconstruction within the POCS framework becomes
	\begin{equation}
	\imb^{t+1} = \mathcal{P}_{DC}  \mathcal{P}_{prior} \imb^t . 
	\end{equation}

	\revb{The prior term in the DDP method does not explicitly provide information on the phase, therefore, reconstruction of the phase is driven by the data consistency projection in the update equation above. For acquisition with multiple coils, the reconstruction method recovers the phase without any modification. For single coil acquisitions, to account for the less amount of information we use an additional projection $\mathcal{P}_{phase}$ in the update equation. A reconstruction iteration is then given as $ \imb^{t+1} = \mathcal{P}_{DC} \mathcal{P}_{phase}  \mathcal{P}_{prior} \imb^t$. We use the $||C \exp(i \angle \imb) ||$ term~\cite{zhaophase} as the regularization with $C$ as the finite difference operator, to prefer smooth phase images. We implement $\mathcal{P}_{phase}$ as taking 10 steps with a step size of 0.1 in the negative gradient direction of the regularization term. It is however possible to change this to other constraints on the phase depending on the application, such as a zero-divergence constraint for phase contrast flow imaging reconstruction~\cite{Santelli}. It is also possible to extend the VAE model and train a prior for complex image patches with an appropriate training set.}

	We apply $T$ POCS steps to complete the reconstruction. Algorithm~\ref{alg:recon} provides a summary of the reconstruction procedure.

\subsection{Experimental setup}\label{sec:setup}
	\subsubsection{MR image data}
	We used structural images from three different data sources to demonstrate the proposed algorithm. 

	First, we used images from the Human Connectome Project (HCP) data set~\cite{VANESSEN20122222} (see https://www.humanconnectome.org/study/hcp-young-adult/document/500-subjects-data-release). The high quality and large number of images from the HCP dataset are ideal for learning priors with the VAE model. 
	We took 2D slices from the T1 weighted 3D MPRAGE images from 158 subjects (790 images in total) to train the prior VAE model. We normalized the training images by mapping their 99$^{th}$ intensity percentile to 1 per image slice. The VAE model was trained for 200k iterations with a batch size of 50.

	Second, to verify that the proposed reconstruction method can be used on a domain that is different from the one the prior is trained on, we experimented with two slices from the Alzheimer's Disease Neuroimaging Initiative (ADNI) data set (for up-to-date information, see www.adni-info.org). 
	The images were selected from subjects with Alzheimer's disease and who have visible white matter lesions. 
	Images with lesions allowed us to also test whether the proposed method will be able to faithfully reconstruct such lesions. We extracted the central slices that showed the largest lesions from these images and further cropped the FOV to 168x224 to remove the empty regions in the images to accelerate computations. 

	\revb{Lastly, we acquired images of 8 healthy volunteers after written informed consent and according to the applicable ethics approval, using a similar protocol to the HCP and ADNI datasets with a 3T Philips Ingenia scanner. 
	The FOV was planned according to the volunteer and a head coil array with 16 elements for receiving was used. 
	We acquired fully sampled complex k-space data for retrospective undersampling study, measuring the coil sensitivities using the standard Philips SENSE reference scan (for only 6 volunteers, due to technical issues) and also estimating them using ESPIRiT~\cite{espirit} (Code: http://people.eecs.berkeley.edu/$\sim$mlustig/Software.html) with default parameters to obtain the autocalibrated coil maps.  }

	\subsubsection{Setup and evaluation}
	We used images from 17 HCP subjects (separate from training subjects), the ADNI images and the acquired data from 8 subjects in our evaluation. 
	We retrospectively undersampled the test images in k-space, reconstructed them back and compared the results with the original images. 
	For the acquired data, the fully sampled k-space data was present while for HCP and ADNI images, we used Fourier transform to compute the fully sampled k-space data.
	We experimented with varying undersampling (US) ratios, which we denote with R when presenting results. 

	We used Cartesian US in one dimension while fully sampling in the other dimension, corresponding to phase-encoding and readout directions, respectively. We present an example of US patterns in Figure~\ref{fig:recres_r3}. We generated the patterns by randomly sampling a 1D Gaussian distribution along the phase-encoding dimension. We randomly drew many sampling patterns from the Gaussian distribution and selected the ones with the best peak-to-side ratio. In addition, we added the central 15 profiles to these selected patterns to fully sample the low-frequency components. We used 2, 3, 4 and 5 for net US ratios (including the fully sampled center). In addition, we used 4-fold radial US with non-uniform fast Fourier transform (NUFFT)~\cite{fesslerNUFFT}. We used the implementation given in~\cite{pyNUFFT} with a conjugate gradient solver. For reconstruction in this case, we only changed the fast Fourier transform (FFT) to NUFFT in the data consistency term and did not modify the prior projection.

	At reconstruction time the undersampled images need to be in the intensity range for which the DDP prior is trained. To meet this requirement, we normalized the undersampled images to their $99^{th}$ percentile before reconstruction with DDP. Notice that this may not be the same value as the $99^{th}$ percentile of the fully sampled images, which would not be available in a real setting for test images. 

	While assessing our proposed DDP method, we generated a new random US pattern for each test image to make sure our empirical analyses assess the effects of the variability of the US patterns.  We reconstructed the test images from HCP and ADNI using 30 POCS iterations (T=30) for R=2 and 3 and 60 for R=4 and 5, 10 iterations for the prior projection (K=10) and the step size was set to $\alpha$=1e-4, where convergence was observed. \revb{We initialize the POCS algorithm using the zero-filled image. When reconstructing images from volunteers, we run 10 iterations of only data consistency projections initially and then turn on the prior projection. We also use lower number of iterations (T=5 for both R=2 and R=3) to prevent divergence due to discrepancy between the true and used coil sensitivities. We implemented the coil combination as given in~\cite{coilcomb} at each data projection step.} 

	We compared the reconstruction with the fully sampled "ground truth" images using Root-Mean-Squared-Error (RMSE), Contrast-to-Noise-Ratio (CNR) and Contrast difference (CN) computed at the gray and white matter boundary in the brain. We use the FreeSurfer~\cite{freesurfer} to obtain the segmentations required for CNR and CN. Details are given in the supplementary materials. We present these results in Section~\ref{sec:results}. 

	\reva{Additionally, we performed three experiments to test sensitivity of ADMM-Net to deviations in acquisition specifications between undersampled images used in training and test. Experiments in \cite{automap} provide evidence to this end for basic feed-forward networks. Authors showed that when k-space trajectories between training and test images differ, the feed-forward network's performance decreased. ADMM-Net is notably different than a basic feed-forward network as it integrates explicit data consistency in the feed-forward architecture. In these experiments we test the null hypothesis that using a different US pattern in reconstruction than in training does not result in a decrease in the performance in the RMSE sense. When measuring RMSE, we used a mask to only measure the reconstruction errors in the brain tissue and skull ignoring artifacts in the background. In each experiment, we trained two networks using different US patterns and applied them on the same test images. In Experiment I, we used R=2 and R=4 for training and tested on R=2. In Experiment II, we used the same networks as in I but tested on R=4. Lastly, we trained networks with R=3 Cartesian and R=3 pseudo-radial patterns and tested on R=3 Cartesian. We used the 17 test images for the evaluation and performed paired Wilcoxon signed-rank tests to assess the null hypothesis.}

	We further experimented with different latent space dimensions, patch sizes and signal-to-noise ratio in the measurements to better characterize the proposed algorithm. Due to space restrictions, we present these analyses in the Appendix.

	\subsubsection{Compared methods}
	We implemented several methods to compare against our proposed approach. These are zero-filling reconstruction, total variation (TV)~\cite{Lustig2007}, dictionary learning (DLMRI)~\cite{dlrecon}, ADMM-Net~\cite{NIPS2016_6406} ,BM3D-MRI~\cite{Eksioglu2016}, SIDWT~\cite{ning_sidwt}, FDLCP~\cite{zhan_fdlcp} and PBDW~\cite{qu_pbdw}. Available implementations for the last three methods only ran on cropped FOV images (square image slices). To compare we used a corresponding DDP reconstruction on cropped FOV. We give the implementation details in the supplementary materials.

\section{Results}\label{sec:results}
	\begin{figure}
	\includegraphics[width=\textwidth]{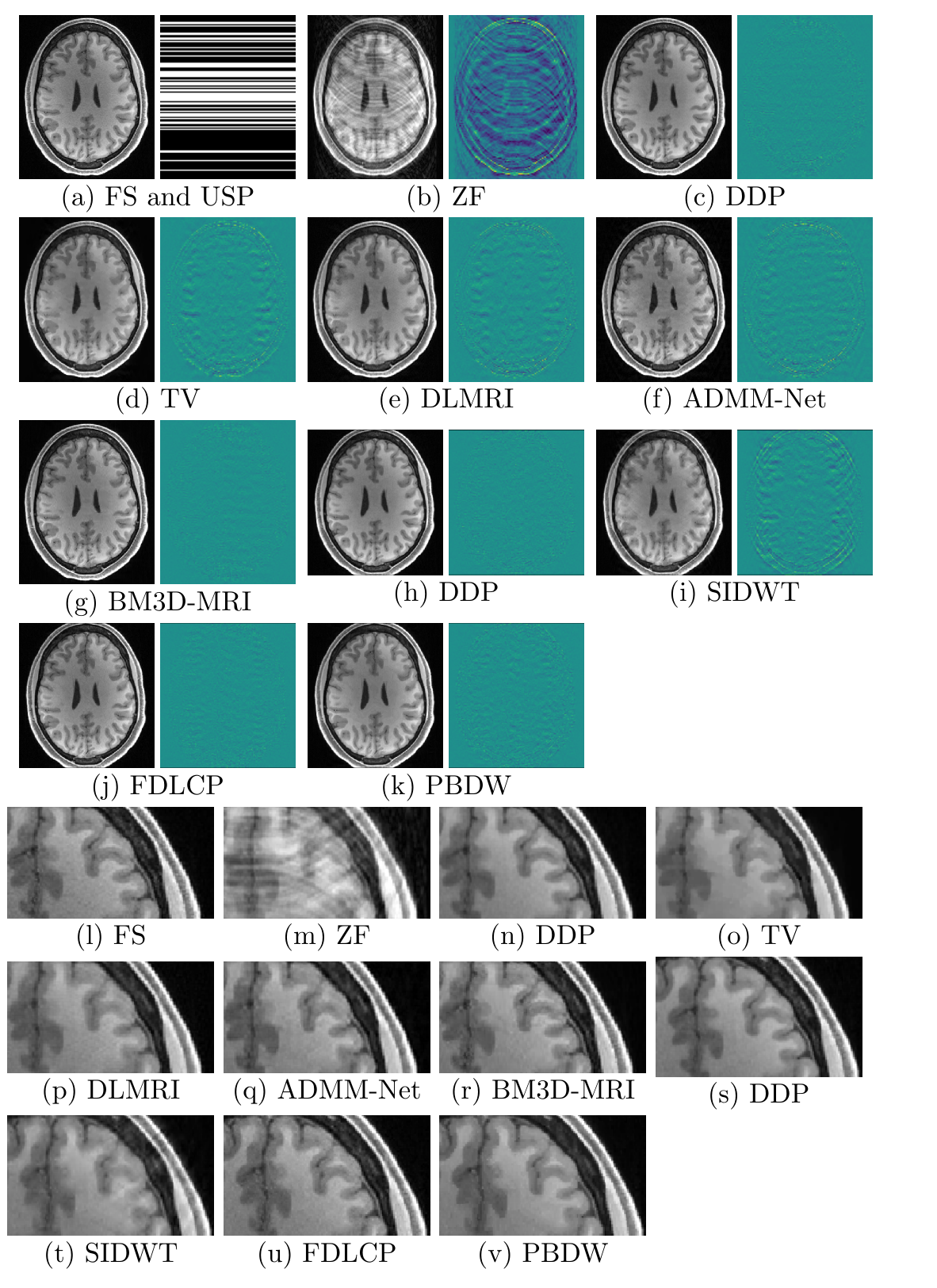}
	\caption{Reconstruction results for R=3. (a) shows the fully sampled image (FS) and the undersampling pattern (USP). (b-g) show reconstruction results and the error maps (intensities clipped to (-0.3, 0.3)) for the full FOV. (h-k) show the results for the cropped FOV. (l-v) show a zoomed region from the images above. (b-g) are produced with the undersampling pattern that was used to train the ADMM-Net for comparability. In (h-k), we also used the same pattern for undersampling for all methods for comparability.}
	\label{fig:recres_r3}
	\end{figure}

	We start by showing visual reconstruction results. Figure~\ref{fig:recres_r3} shows results for one of the test images from the HCP data for R=3. The sampling pattern is also shown in the figure. Visual quality of reconstructions for different methods varied. TV and PBDW reconstructions had problems in restoring the structure and texture. Respective reconstructed images in Figure~\ref{fig:recres_r3} appear cartoon-like and error images show higher error in regions where gray (GM) and white matter (WM) structures are intertwined. SIDWT did not complete dealiasing and ADMM-Net did not reconstruct the small GM island in the zoomed image. DDP, DLMRI and FDLCP perform well. We show four more randomly selected images from the test set in supplementary Figures S6 and S7 for R=2,3 as well as reconstruction results for R=2,4,5 in supplementary Figures S8, S9 and S10.

	We present the quantitative results for reconstruction accuracy in Table~\ref{table:errors_new}. We show results for full FOV and cropped images separately as the numbers are not comparable. In terms of RMSE our DDP method performed best for the full FOV reconstructions (except for R=2, where BM3D-MRI performed better) and second best against FDLCP for the cropped FOV setting. In terms of CNR, the proposed method performed equally or better than all the other methods for both full and cropped FOVs. 

\begin{figure}
	\includegraphics[width=\textwidth]{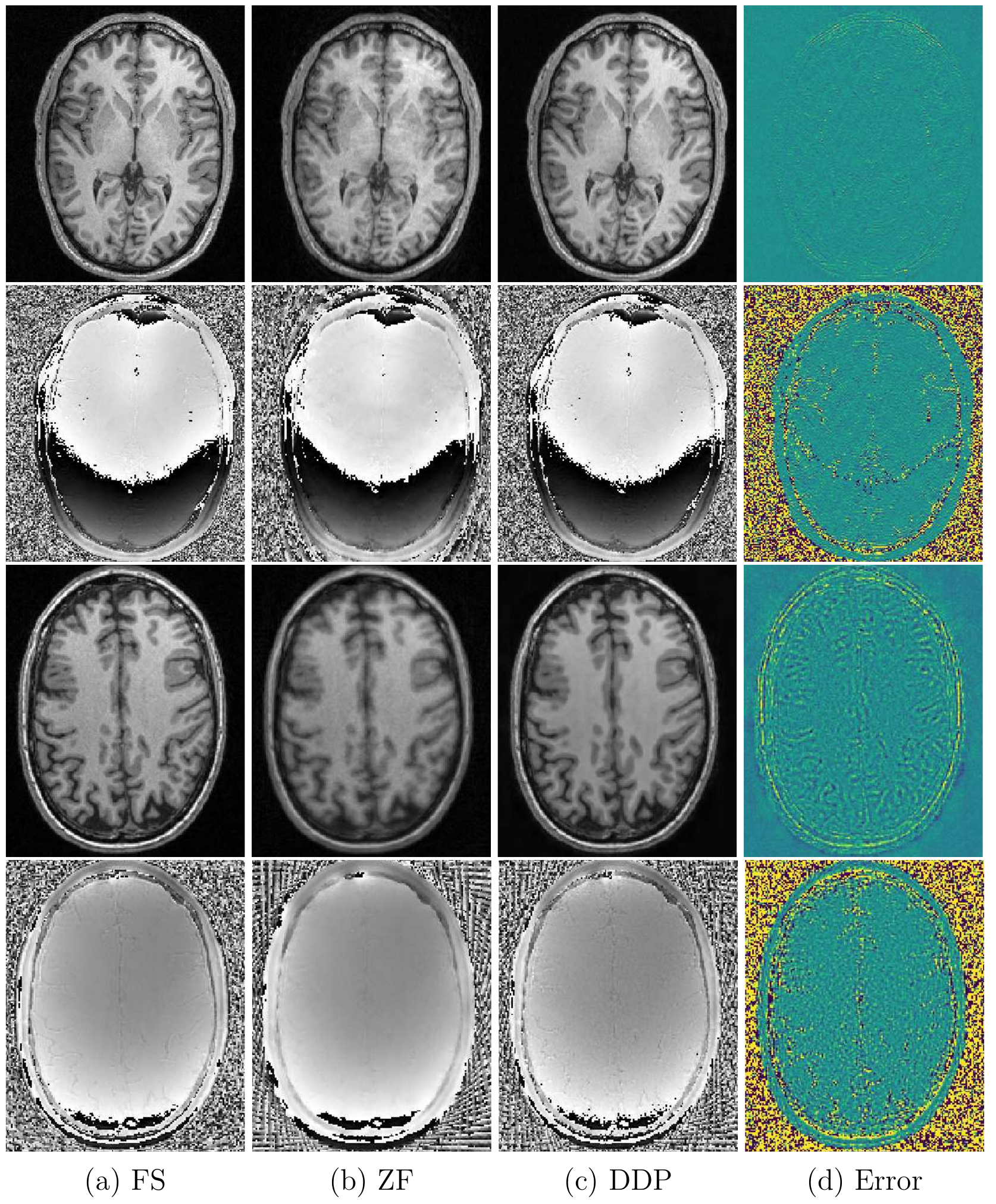}
	\vspace{-0.5cm}
	\caption{Reconstruction results for measured data. Rows show the magnitude and phase of (from left to right) the fully sampled image, (FS), the zero-filling image (ZF), the DDP reconstruction (using ESPIRiT coil maps) and the error (clipped to (-0.3,0.3)). Upper two rows: A Cartesian US pattern (R=2) was used to retrospectively undersample the k-space data. Lower two rows: A radial US pattern (R=4) was used. In both cases the prior projection is the same, only the data consistency projection differs (using FFT and NUFFT, respectively).} 
	\label{fig:recres_real_data}
	\end{figure}

	\revb{In Figure~\ref{fig:recres_real_data} we show reconstruction results from retrospectively undersampled k-space data acquired within this study. First example is reconstruction of 2-fold Cartesian undersampling and the second is of 4-fold radial undersampling using NUFFT in the encoding operation. Our DDP method can reconstruct the original magnitude and phase fairly well, though the magnitude image was smoother than the fully sampled magnitude image in both cases. Quantitatively, the mean (std) RMSE for the Cartesian case for all subjects was 6.97\% (0.37) using ESPIRiT coil maps (N=8) and 6.92\% (0.58) using measured coil maps (N=6) for R=2. Mean (std) value for R=3 was 10.35\% (1.53) with ESPIRiT coil maps and 9.93\% (1.82) with measured sensitivities. These values were higher than those for HCP data. In order to see the portion of the increase in RMSE due to the inaccuracies in the encoding operator and phase, we also reconstructed the images by setting the phase of the fully sampled image to zero before undersampling and expanding the image by known coil maps. In this case the mean RMSE reduced to 5.76\% (0.56) (R=2, N=8). Furthermore, we also reconstructed the images using only data consistency projections, corresponding to SENSE reconstruction, to see the added value of the DDP projections. In this case the mean RMSE was 7.95\% and 11.42\% for (R=2,3, respectively, N=8, ESPIRiT maps). We present more information on this point in the supplementary materials.}    

	\begin{table*}[bt]
	\caption{Table summarizing results for different reconstruction quality metrics. Numbers indicate the mean (and standard deviation) of the error metric for N=17 test images. Top group are the results for experiments with full FOV images and bottom group are for cropped FOV images.}
	\resizebox{\textwidth}{!}{
	\begin{tabular}{l|l|l|l|l|l|l|l|l|l|l|l|l|}
	\cline{2-13}
	                                & \multicolumn{3}{l|}{R=2} & \multicolumn{3}{l|}{R=3} & \multicolumn{3}{l|}{R=4} & \multicolumn{3}{l|}{R=5} \\ \cline{2-13} 
	                                & RMSE    & CNR    & CN    & RMSE    & CNR    & CN    & RMSE    & CNR    & CN    & RMSE    & CNR    & CN    \\ \hline
	 \multicolumn{1}{|l|}{FS}  &  -  & 0.48(0.10) & 0.12(0.02) & - & 0.48(0.10) & 0.12(0.02) & - & 0.48(0.10) & 0.12(0.02)& - & 0.48(0.10) & 0.12(0.02) \\ \hline
	 \multicolumn{1}{|l|}{Zero-fill} &  13.03(1.13) & 0.40(0.09) & 0.12(0.02) & 21.15(1.36) & 0.33(0.07) & 0.09(0.02)  & 24.92(1.91) & 0.31(0.06) & 0.08(0.02)& 27.36(1.79) & 0.30(0.06) & 0.08(0.02)\\ \hline
	 \multicolumn{1}{|l|}{DDP}  & 2.76(0.53) & 0.48(0.11) & 0.12(0.02) & 4.25(0.61) & 0.48(0.10) & 0.12(0.02) & 6.46(1.57) & 0.46(0.11) & 0.11(0.02)& 11.13(2.39) & 0.41(0.10) & 0.10(0.02) \\ \hline
	 \multicolumn{1}{|l|}{TV}  & 3.87(0.47) & 0.46(0.11) & 0.12(0.02) & 7.56(0.83) & 0.40(0.10) & 0.09(0.02) & 11.40(1.39) & 0.35(0.09) & 0.08(0.02)& 14.56(1.23) & 0.31(0.08) & 0.07(0.02) \\ \hline
	 \multicolumn{1}{|l|}{DLMRI} &  4.48(0.52) & 0.46(0.11) & 0.12(0.02) & 7.25(0.83) & 0.40(0.10) & 0.10(0.02) & 10.72(1.31) & 0.33(0.09) & 0.08(0.02)& 13.87(1.25) & 0.30(0.08) & 0.07(0.02) \\ \hline
	 \multicolumn{1}{|l|}{ADMMNet}  & 3.55(0.40) & 0.48(0.11) & 0.12(0.02) & 7.06(0.52) & 0.45(0.11) & 0.11(0.02) & 11.26(0.72) & 0.36(0.09) & 0.09(0.02)& 13.05(0.70) & 0.32(0.08) & 0.08(0.02) \\ \hline
	 \multicolumn{1}{|l|}{BM3D-MRI} &  1.92(0.36) & 0.48(0.10) & 0.12(0.02) & 4.23(1.05) & 0.46(0.10) & 0.11(0.02) & 8.08(2.48) & 0.43(0.10) & 0.10(0.02)& 11.70(2.76) & 0.38(0.09) & 0.09(0.02) \\ \hline \hline
	 \multicolumn{1}{|l|}{FS}  & - & 0.48(0.10) & 0.12(0.02) & - & 0.48(0.10) & 0.12(0.02) & - & 0.48(0.10) & 0.12(0.02)& - & 0.48(0.10) & 0.12(0.02) \\ \hline 
	 \multicolumn{1}{|l|}{DDP} &  2.68(0.38) & 0.48(0.10) & 0.12(0.02) & 4.61(1.12) & 0.47(0.10) & 0.11(0.02) & 7.39(1.47) & 0.45(0.10) & 0.11(0.02)& 13.00(3.01) & 0.39(0.08) & 0.10(0.02) \\ \hline
	 \multicolumn{1}{|l|}{SIDWT} &  4.49(0.98) & 0.45(0.11) & 0.12(0.02) & 9.42(1.62) & 0.39(0.09) & 0.12(0.02) & 14.57(1.96) & 0.33(0.08) & 0.10(0.02)& 18.76(2.80) & 0.32(0.07) & 0.10(0.02) \\ \hline
	 \multicolumn{1}{|l|}{FDLCP} &  2.63(0.35) & 0.48(0.10) & 0.12(0.02) & 4.35(0.87) & 0.45(0.10) & 0.14(0.03) & 6.72(0.89) & 0.41(0.10) & 0.13(0.03)& 9.62(1.48) & 0.35(0.08) & 0.11(0.02) \\ \hline
	 \multicolumn{1}{|l|}{PBDW}  & 3.24(0.38) & 0.47(0.11) & 0.12(0.02) & 5.59(0.94) & 0.44(0.10) & 0.13(0.02) & 8.51(0.98) & 0.38(0.09) & 0.12(0.02)& 11.38(1.39) & 0.34(0.08) & 0.10(0.02) \\ \hline
	\end{tabular}}
	\label{table:errors_new}
	\end{table*}

\begin{figure}
	\includegraphics[width=\textwidth]{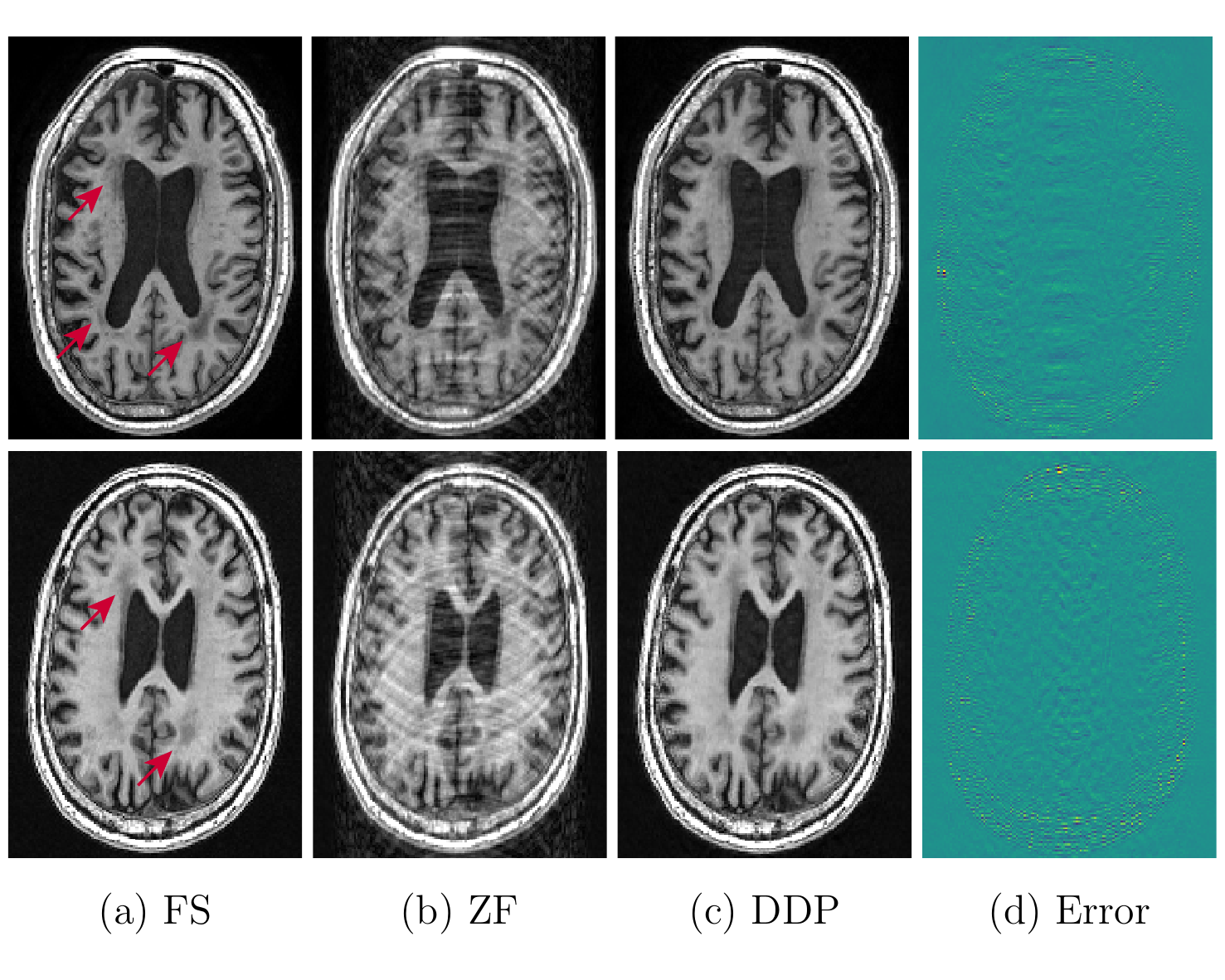}
	\caption{DDP reconstruction results for two images with white matter lesions due to Alzheimer's disease from the ADNI data set for R=2. Images show the original (FS), zero-filling (ZF), reconstructed (DDP) and the error maps from left to right. Lesions are clearly visible in the reconstructed images as well.  Error map values are clipped to (-0.3,0.3). Arrows denote lesions.}
	\label{fig:recadni}
	\end{figure}

	In Figure~\ref{fig:recadni}, we show DDP reconstructions for the ADNI images for R=2. We used the VAE model that was trained on the HCP dataset, which had only healthy subjects, to reconstruct the images here. The reconstructed images recover GM and WM structures and edges faithfully. The WM lesions were also well reconstructed. The error maps do not indicate a specific increase in the lesion regions. 

	In addition to these results, we show the convergence of the POCS algorithm in Figure S3, and results for patch size, latent dimension and noise analyses in Figures S4 and S5 of the supplementary materials. 

	\begin{table}[]
	\resizebox{\textwidth}{!}{
        		\begin{tabular}{l|l|l|l|l|l|l|}
\cline{2-7}
                                          & \multicolumn{2}{l|}{Experiment I} & \multicolumn{2}{l|}{Experiment II} & \multicolumn{2}{l|}{Experiment III} \\ \hline
\multicolumn{1}{|l|}{R\_train}            & 2               & 4               & 4                & 2               & 3 cart           & 3 rad            \\ \hline
\multicolumn{1}{|l|}{R\_US}               & 2               & 2               & 4                & 4               & 3 cart           & 3 cart           \\ \hline
\multicolumn{1}{|l|}{RMSE mean}           & 3.17            & 4.11            & 10.55            & 10.70           & 6.49             & 6.67             \\ \hline
\multicolumn{1}{|l|}{RMSE std}            & 0.39            & 0.44            & 0.68             & 0.66            & 0.49             & 0.51             \\ \hline
\multicolumn{1}{|l|}{RMSE diff. 25\%}     & \multicolumn{2}{c|}{0.82}         & \multicolumn{2}{c|}{0.09}          & \multicolumn{2}{c|}{0.14}           \\ \hline
\multicolumn{1}{|l|}{RMSE diff. median}   & \multicolumn{2}{c|}{0.86}         & \multicolumn{2}{c|}{0.15}          & \multicolumn{2}{c|}{0.16}           \\ \hline
\multicolumn{1}{|l|}{RMSE diff. 75\%}     & \multicolumn{2}{c|}{1.02}         & \multicolumn{2}{c|}{0.22}          & \multicolumn{2}{c|}{0.2}            \\ \hline
\multicolumn{1}{|l|}{p-value}             & \multicolumn{2}{c|}{0.0002}       & \multicolumn{2}{c|}{0.0009}        & \multicolumn{2}{c|}{0.0002}         \\ \hline \hline
\multicolumn{1}{|l|}{RMSE DDP mean (std)} & \multicolumn{2}{l|}{2.47 (0.44)}  & \multicolumn{2}{l|}{6.08 (1.45)}   & \multicolumn{2}{l|}{3.93 (0.56)}   \\ \hline
\end{tabular}}
	\caption{\reva{Reconstruction results for ADMM-Net, when using different US patterns at training and test times. Statistics were derived from N=17. R$_{train}$ and R$_{test}$ denote the US factor (pattern) used in training and testing. If not stated, a Cartesian pattern (cart) was used and "rad" denotes pseudo-radial pattern. RMSE diff. refers to the n$^{th}$ percentile or median value of the distribution of pairwise RMSE differences. p-values are calculated with paired Wilcoxon signed-rank test. Last row presents the DDP RMSE values for comparison purposes. These RMSE values were calculated with disregarding the artifacts in the background using a brain mask. }}

	\label{table:admmnet_stuff}
	\end{table}

	\reva{Lastly, we show results for the experiments assessing sensitivity to deviations in acquisition specifications between training and test images of ADMM-Net in Table~\ref{table:admmnet_stuff}. All the differences were statistically significant at the 0.05 level according to the paired test, performance of the method decreased significantly when the training and test patterns/ratios differed. Performance differences in the Experiment I were particularly high. }
\section{Discussion}
	The reconstruction examples in Figure~\ref{fig:recres_r3} and the quantitative results in Table~\ref{table:errors_new} show that the proposed deep density prior reconstruction method produced highly accurate reconstructions both in terms of RMSE and CNR compared to the other methods. While BM3D-MRI and FDLCP could achieve slightly lower RMSE's, they both decreased CNR further than DDP, indicating that both methods sacrificed contrast to reduce RMSE, generating blurrier images. Leveraging the powerful prior over image patches, DDP was able to restore structures in the image faithfully while removing aliasing artifacts. Comparing methods that use explicit priors, DDP's performance demonstrate the huge potential in DL-based priors for MRI reconstruction. Compared to ADMM-Net, DDP's higher accuracy show that it is a highly attractive alternative to feed-forward network approaches. 

	Experiments with images from the ADNI dataset and acquired k-space data showed further properties of DDP and also prompted some key questions for further research. \revb{Results in Figure~\ref{fig:recres_real_data} show that the DDP method yielded high quality reconstructions for both the magnitude and phase images, though with higher mean RMSE values than those in the HCP test images. The fact that the RMSE decreased when the correct coil sensitivities and phase were used, suggest that some part of the error is due to the discrepancies between real and measured coil sensitivities. Hence, increasing accuracy of the coil maps would potentially decrease the error rate. Similarly, the method currently uses theoretical or no prior for the phase but it would benefit from better priors for this component as well.}

	The ADNI reconstruction results are encouraging in two respects. First, they show that the learned model does not blur out the lesions during the prior projection, which could have been expected since the training images did not contain any examples of lesions. \revc{We believe the lesions could be reconstructed because they are structurally not very different from healthy brain structures. In addition, the data consistency term makes sure the sampled information regarding the lesions is kept in the final image.} \revc{However, the proposed method, similar to other reconstruction methods, requires further investigation as to where its limits lie, such as with bigger, brighter lesions.} We believe for an optimal treatment of lesions, the training data set should include such images to provide the prior the capability to represent them.

	Second, the proposed method performed reasonably well despite the domain difference between the training and test sets. Although the two data sets were acquired at the same field strength, they still differed in some acquisition protocol and imaging parameters. Their FOV and voxel resolution are different, and HCP was acquired with fat suppression while ADNI was not, which affects the image contrast. Our method is invariant to changes in FOV but not in scale. The lack of fat suppression also makes the dealiasing more challenging in the ADNI images \reva{since artifacts become more prominent}. This results in very faint artifacts in the ventricles in the reconstructed images. Despite these differences, the quality of the reconstruction results indicate that the learned prior model could generalize to slightly different scales and similar but not identical imaging protocols. However, these experiments raised two key questions and exciting research directions regarding the sensitivity of the proposed method. 

	\reva{Even though the proposed method is not sensitive to variations in undersampling patterns or coil settings, when contrast or resolution of the acquired images differ substantially from training images used to learn the prior, we believe reconstruction quality will decrease. There are two interesting directions to remedy this issue. First is to use a different prior for different contrast and resolutions. Second is to further improve the reconstruction quality using appropriate domain adaptation methods. Integrating invariance to domain differences in the prior and even building a joint prior for multiple contrasts are interesting research directions. Similar sensitivity issues can also be expected when the underlying anatomy differs between the acquired image and the prior. In this scenario, the safest approach would be to train a prior for each anatomy but building a prior that is capable of representing multiple anatomies is also an interesting direction. We also note that it is possible that all learning based methods suffer from the mentioned sensitivities~\cite{domadapt}.}

	An underlying assumption of our model is the unit Gaussian prior for the latent space, for which different extensions such as Gaussian mixtures or graphical models were investigated in~\cite{gmvae, graphvae},  which could in principle improve the representation capacity of the prior. Similarly, different density estimation methods are investigated, for instance using generative adversarial networks (GAN)~\cite{gan, 8237889}. \revb{In~\cite{ledig}, the authors proposed to use the discriminator of a GAN to superresolve images. The authors in \cite{shah} explain the drawbacks of this approach and similar to \cite{yeh} suggest projecting images onto the span of the generator. A similar approach is given in }\revb{\cite{mardani}. The advantage of using VAEs is the simplicity of the inverse mapping. Determining the latent space representation of a given image is much more straightforward with VAEs, not requiring a separate optimization as GANs. }

	\revb{Another alternative approach to the VAE would be to use denoising autoencoders (DAE)~\cite{denae}. Even though these might offer faster algorithms, the \revb{advantages} of VAEs compared to DAEs are i) a more principled way of approximating the target distribution and ii) ease and rigor in approximating the likelihood of an image patch due to the variational inference mechanism that underlies VAEs.}

	One limitation of our method is the requirement of training in contrast to methods using fixed bases for regularization. Training of the prior relies on the availability of high quality data. Such data is available for commonly used sequences, such as structural T1w and T2w MRI. For other sequences, such as functional or diffusion MRI, construction of appropriate training set is of interest for future research.

	Experiments with different configurations, which are presented in the supplementary materials, showed that the DDP reconstruction is not sensitive to patch size and latent space dimensions for a reasonable range. Furthermore, reconstruction quality is higher when patches larger than 12x12 are used. In addition, as expected, the performance degrades with decreasing SNR, however, with a low rate indicating robustness of the method to noise. \revc{This means the method is likely to perform fairly well in the regions with higher g-factor that have lower SNR due to parallel imaging.}

	\revb{In our current implementation a single evaluation of the derivative through the network takes around 9 seconds for 360 28x28 patches (0.6 seconds per batch of 25 patches in parallel). The total run time is given by this multiplied by number of iterations and all the other operations including mainly the phase projection, data projection and application of the derivatives for these operations. This time can be reduced by optimizing the code, increasing the parallelization and changing the network to work with images directly rather than patches.} 
	\reva{Furthermore, we demonstrate our method on 2D slices with a single phase encoding/undersampling direction as a proof of concept. However the method can be extended to a 3D setting, either by doing slice-by-slice reconstruction with a VAE trained on the whole brain or by training a VAE with 3D patches.}


	\reva{Results presented in Table~\ref{table:admmnet_stuff} demonstrate that ADMM Net model is also sensitive to deviations in undersampling patterns used in training and test images, similar to the method proposed in~\cite{automap}. Other feed-forward networks may also show similar sensitivities however, to the best of our knowledge, such an analysis is rarely performed.} In contrast, we emphasize that the proposed method, due to decoupling of the prior and the data consistency term, does not share the same sensitivity. It learns the prior distribution on fully-sampled images and can be used to reconstruct any sampling scheme faithful to the measured data without the need of retraining as long as the images are from the same domain. 
	\section{Conclusion}
	In this paper we proposed a novel method termed DDP for MR reconstruction from undersampled k-space acquisitions. The method uses the VAE algorithm to learn the distribution of MR patches from fully sampled images, removing the sensitivity of the model to the sampling pattern. The model then uses this learned distribution as a probabilistic prior in a Bayesian reconstruction framework. We have shown that the reconstruction with the DDP approach yielded promising results for HCP and ADNI data sets as well as multi-channel k-space measurements in terms of visual quality and quantitative measures.

\vspace{-0.25cm}
\section*{Acknowledgments}
Authors would like to acknowledge Prof. Sebastian Kozerke for valuable discussions. We thank GyroTools for their MRecon software. We thank NVIDIA for their GPU donation. We also thank the reviewers for comments/suggestions that substantially improved this work. We used images from the HCP and ADNI datasets.

\ifCLASSOPTIONcaptionsoff
  \newpage
\fi



\bibliographystyle{IEEEtran}
\bibliography{library.bib}

\beginsupplement

\section{Supplementary Materials to MR image reconstruction using deep density priors}\label{sec:vae_deatils}
In this accompanying document, we provide supplementary details and results to support the main document. 

\subsection{Details on VAE network architecture and training}
\begin{figure}[!htb]
\begin{center}
\includegraphics[trim={0.2cm 0.5cm 1.5cm 0.3cm},clip,width=0.9\linewidth]{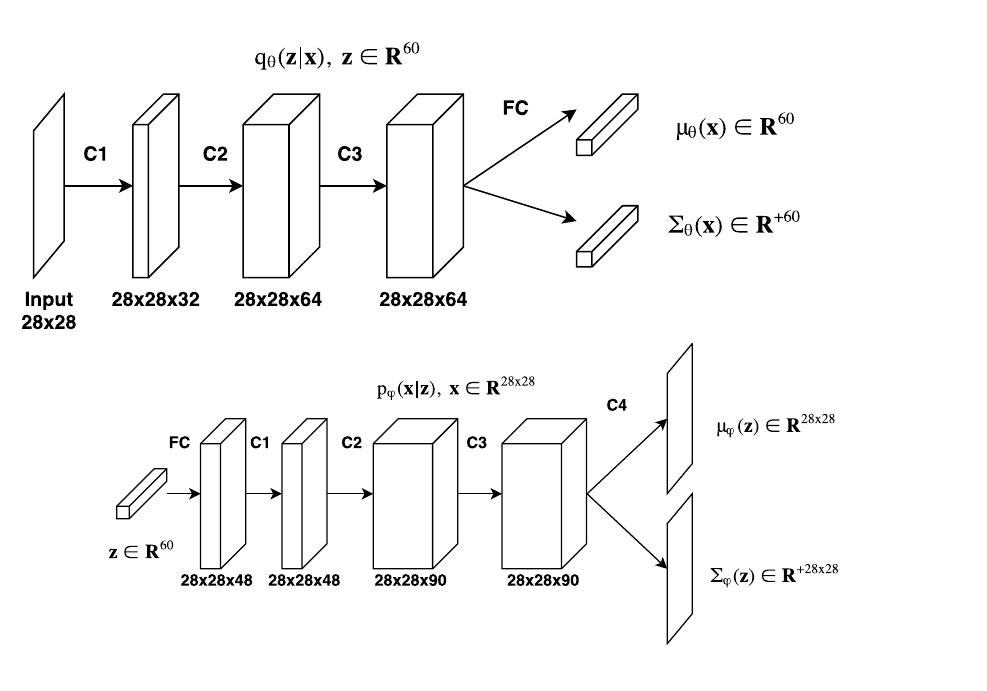}
\end{center}
\caption{\label{fig:networks} Architecture of the encoding (top) and decoding (bottom) networks of our VAE. Arrows indicated with $C\#$ are convolutional layers followed by ReLU non-linear activation except $C4$ of decoding network, which is not followed by non-linear activation. Arrows indicated by FC are fully connected layers. FC of the decoding network is followed by a ReLU activation but not of the encoding network.}
\end{figure}

Both $p_{\varphi}(|\xb||\zb)$ and $q_{\theta}(\zb | |\xb|)$ networks are depicted in Figure~\ref{fig:networks}. For the encoding network, the input is an image patch and the output is the mean and the covariance of the corresponding posterior. The decoding network takes a vector $\zb$ as input and outputs the mean and the covariance of the corresponding likelihood for the patch. Both networks are mostly convolutional with single fully connected layers. All convolutional layers for both networks use 3x3 kernels and all layers also have additive bias terms throughout the network. We use rectified linear units (ReLU) as the non-linear activation function.

To avoid numerical stability issues we use the log of the variance values throughout the network. We initialize the network weights with a truncated normal initializer with standard deviation 0.05. We use Adam~\cite{adam} for optimization (learning rate of 5e-4, default momentum values in Tensorflow).

As the base model, we used patches of size 28$\times$28 and a 60 dimensional latent space. However, we also experimented with different patch size and latent dimensions. 
\subsection{Details on acquisition parameters of used images}
Images from HCP and ADNI dataset were publicly available. Here, we summarize the acquisition parameters detailed in the respective websites for completeness. 

The HCP images were acquired at a 3T Siemens device with 2400 ms, 2.14 ms and 1000 ms for TR, TE and TI, respectively, flip angle of 8 degrees, a standard field of view for all subjects of 224x224x224 mm$^3$ with 0.7 mm isotropic resolution using a bandwidth of 210 Hz/voxel. Fat suppression was used during acquisition. We used the minimally preprocessed images in order to also have the corresponding FreeSurfer~\cite{freesurfer} segmentations, which we use in our evaluation to compute contrast-to-noise ratio. The preprocessing steps consisted of resampling to a FOV matrix 260x311x260 and rigid alignment. Despite the rigid alignment, there were substantial local orientation differences between the images from different subjects. We used five central slices of the volumes, skipping four slices between each. We cropped the images to a size of 252x308, removing only background, to reduce computational load.

The acquisition parameters of the ADNI images were different than HCP: TR, TE, TI values were 7.34 ms, 3.03 ms, 400 ms, flip angle was 11 degrees, FOV matrix was 196x256x256 with resolution 1.2 mm x 1 mm x 1 mm. The images were acquired with a 3T GE scanner with no fat suppression and were bias-field corrected with the N3 algorithm~\cite{n3paper}. 

Lastly, for the images that were acquired within this study, we used a turbo field echo sequence with 1 mm isotropic resolution, flip angle of 8 degrees, TR and TE set to 8.1 ms and 3.7 ms, an inversion pulse with 1 s delay and a linear sampling order. Images were acquired at 3T with a Philips Ingenia device.

\subsection{Implementation details of the methods used for comparisons}
As the baseline, we evaluated zero-filling (ZF) reconstructions. This baseline represents a lower-bound of the reconstruction accuracy. The first method we used is the total variation (TV) reconstruction as described in~\cite{Lustig2007}. We used the BART toolbox implementation that is publicly available~\cite{bart}, specifically the "pics" tool with TV regularization (regularization strength 0.075) and the ADMM parameter $\rho$ as 1. We used 20 conjugate gradient steps and 4500 total iterations ("bart pics -R T:3:0:0.0075 -u1 -C20 -i4500"). The parameters above are chosen by a grid search for the best parameter setting in the RMSE sense. 

As the second method, we used reconstruction using dictionary learning (DLMRI) as proposed in~\cite{dlrecon}\footnote{Code available at http://www.ifp.illinois.edu/$\sim$yoram/DLMRI-Lab/Documentation.html}. We used 200 iterations, a patch size of 36 voxels and 36 dictionary atoms. Furthermore, we set number of signals used for training to 7200 and the overlap stride to 2. K-SVD learning was used with both sparsity and error threshold. The sparsity level was set to 7. The error threshold was set to 0.046 for the first four iterations, then to 0.0322 for the rest of the iterations. We used 15 K-SVD iterations. We chose the parameters as suggested by the authors in the code, but increased the number of iterations.

Third, we used the ADMM-Net\footnote{https://github.com/yangyan92/Deep-ADMM-Net}~\cite{NIPS2016_6406} algorithm, a feed-forward neural network with data consistency term that unrolls ADMM iterations, designed for square images with radial undersampling. We modified the code to work with non-square images and Cartesian undersampling patterns. To train the model, we used the same 790 images that were used to train the VAE model. In order to correctly evaluate the method, we used the same undersampling pattern to train and test the method, except while performing sensitivity tests described in the main text. We normalized the image intensities in the same way as for DDP. We used 15 stages with 8 filters (filter size 3x3), and set padding as 1. We did not use weight decay during training. We set the maximum iteration number to 25 for the L-BFGS algorithm. It trained for maximum number of iterations for R=2 (45 hours), 13 iterations before convergence for R=3 (42 hours), 18 iterations before running out of time (120 hours) for R=4 and maximum number iterations for R=5 on a GPU (GeForce GTX TITAN X). The normalized mean squared errors were 0.078 and 0.035 for R=2, 0.11 and 0.071 for R=3, 0.15 and 0.11 for R=4, and 0.19 and 0.13 for R=5, before and after training, respectively, on the training set. For the radial undersampling patterns (R=3) used in the sensitivity tests, the network was trained for 36 hours before termination after 11 iterations. We took the parameter setting for which the best results were reported in the paper.

We also compared with SIDWT~\cite{ning_sidwt}, FDLCP~\cite{zhan_fdlcp} and PBDW~\cite{qu_pbdw}, which were reported to achieve highly accurate reconstructions. For these methods we had to crop the images to 256x256 and generate new undersampling patterns for this size since the authors implementations worked only with that size\footnote{Implementations from http://csrc.xmu.edu.cn/csg\_publications\_en.html}. We did not modify the code for these methods, as only the binaries were available, and took the parameters as set by the authors in the code. We ran our proposed method on these images as well.

Lastly, we compared to BM3D-MRI\footnote{Code available at http://web.itu.edu.tr/eksioglue/pubs/BM3D\_MRI.htm}~\cite{Eksioglu2016}, a powerful reconstruction method that leverages redundancy in the images, and used the parameters as set by the author in the code.

\subsection{Performance Metrics}\label{sm:metrics}
Here we supply the formulae of the performance metrics used in the presented evaluation.

Normalized Root-Mean-Squared-Error (RMSE) is defined as:
\begin{equation}
\text{RMSE}(gt,rec) = 100\cdot\sqrt{\frac{\sum_i^N (|gt_i|^2 - |rec_i|)^2}{\sum_i^N |gt_i|^2}},
\end{equation}
where $gt$ and $rec$ are the fully sampled and reconstructed images. $|.|$ denotes the magnitude operator. The subscript i goes through all the N voxels in the image. If a mask is used, then only the voxels in the masked region are used. The denominator provides the normalization factor.

Contrast-to-Noise Ratio (CNR) is given as: 
\begin{equation}
\text{CNR}(rec) = \frac{|mean_{GM}(rec) -  mean_{WM}(rec)|} { std_{GM}(rec) + std_{WM}(rec) },
\end{equation}
where $mean_{GM}$ and $mean_{WM}$ denote the mean values of the voxels in the gray matter and white matter, and $std_{xx}$ denotes the standard deviation in the respective tissue type. To compute these statistics, we use the whole GM but only the edge of the WM. In order to obtain white matter boundaries we applied binary erosion (with a square structuring element of size 7x7) to the white matter segmentations, as computed by FreeSurfer, and took the difference of the original and eroded segmentations. This gives us the voxels at the boundary of WM and GM but within WM.

Contrast-to-noise is similar, but without the division by the standard deviation:
\begin{equation}
\text{CNR}(rec) = |mean_{GM}(rec) -  mean_{WM}(rec)|.
\end{equation}

\subsection{Complementary Results}\label{sm:results}
Here we present some supplementary results to the ones presented in the main text.

\begin{figure}[!htb]
\includegraphics[trim={0 0 0.3cm 0},clip,width=\textwidth]{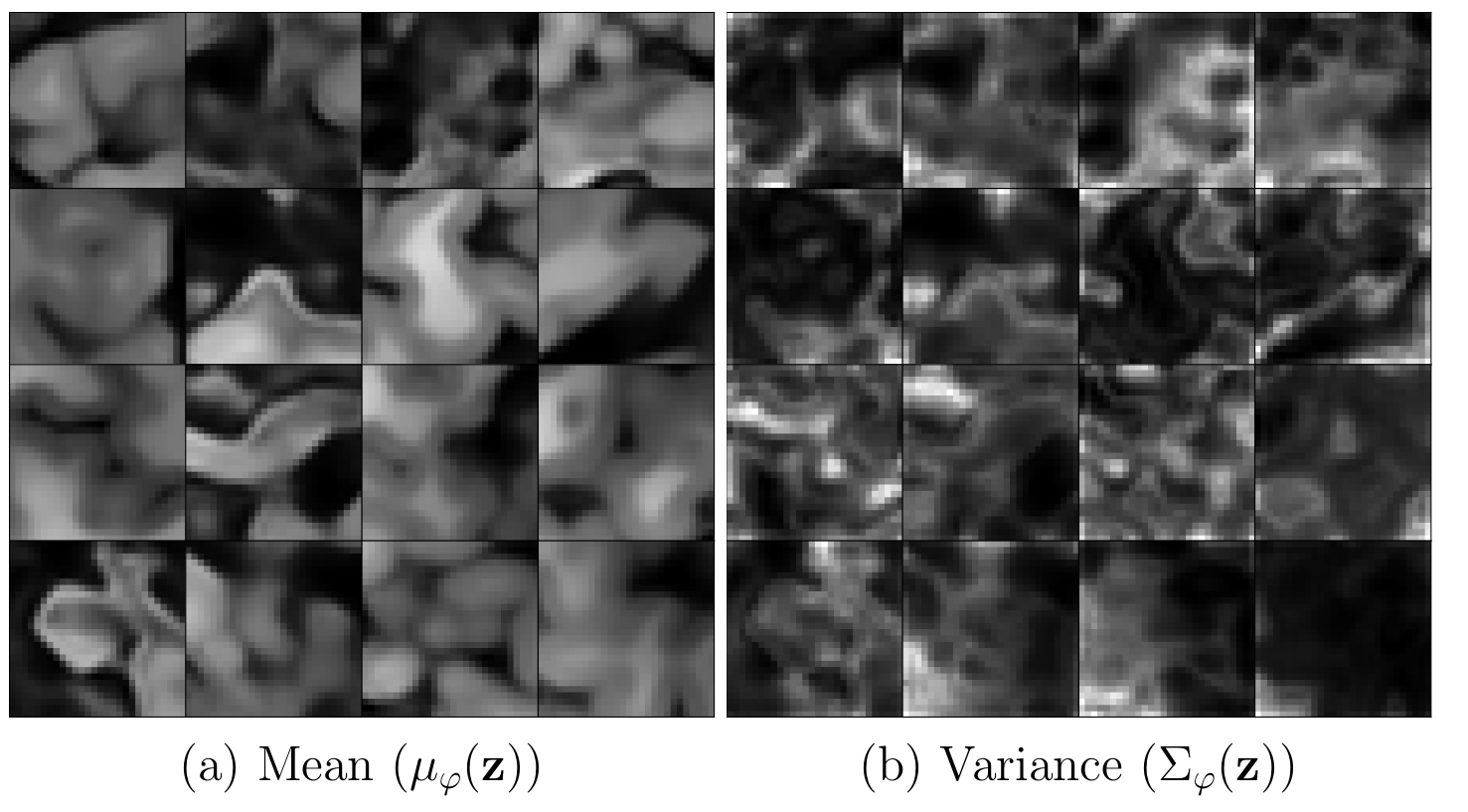}
\caption{16 image patches sampled from the prior learned by the VAE algorithm. Images on the left are the mean predictions and the images on the right are the corresponding variance maps. The patches are 28x28.}
\label{fig:generation}
\end{figure}

Firstly, we show patches sampled from the prior model trained for patch-size of 28x28 and latent dimension of 60 in Figure~\ref{fig:generation}. These patches were generated by simply feeding 16 random $\zb$ vectors drawn from unit Gaussian to the decoder network. The decoder network outputs the mean images, i.e. $\mu_{\varphi}(\zb)$, and the corresponding variance images, i.e. $\Sigma_{\varphi}(\zb)$. We like to note that we have not cherry-picked these examples. The sampled mean patches, shown on the left, look realistic where gray matter (GM), white matter (WM) and gyri/sulci structures are clearly defined and properly positioned. These generated samples suggest that the VAE algorithm is able to approximate the underlying distribution of MR patches. The variance images on the right show that VAE places high variance in the cerebrospinal fluid (CSF) areas as well as boundaries between structures. We observe that isolated GM or CSF islands receive high variance.

The samples support our hypothesis that the VAE model can learn to approximate the distribution of MR patches. The variance maps show that the sulci-like generated structures filled with CSF have higher variance, which is in accordance with the works that quantified uncertainty in image synthesis~\cite{tanno}.

\begin{figure}[ht]
\includegraphics[trim={3cm 9cm 3cm 8cm},clip,width=\textwidth]{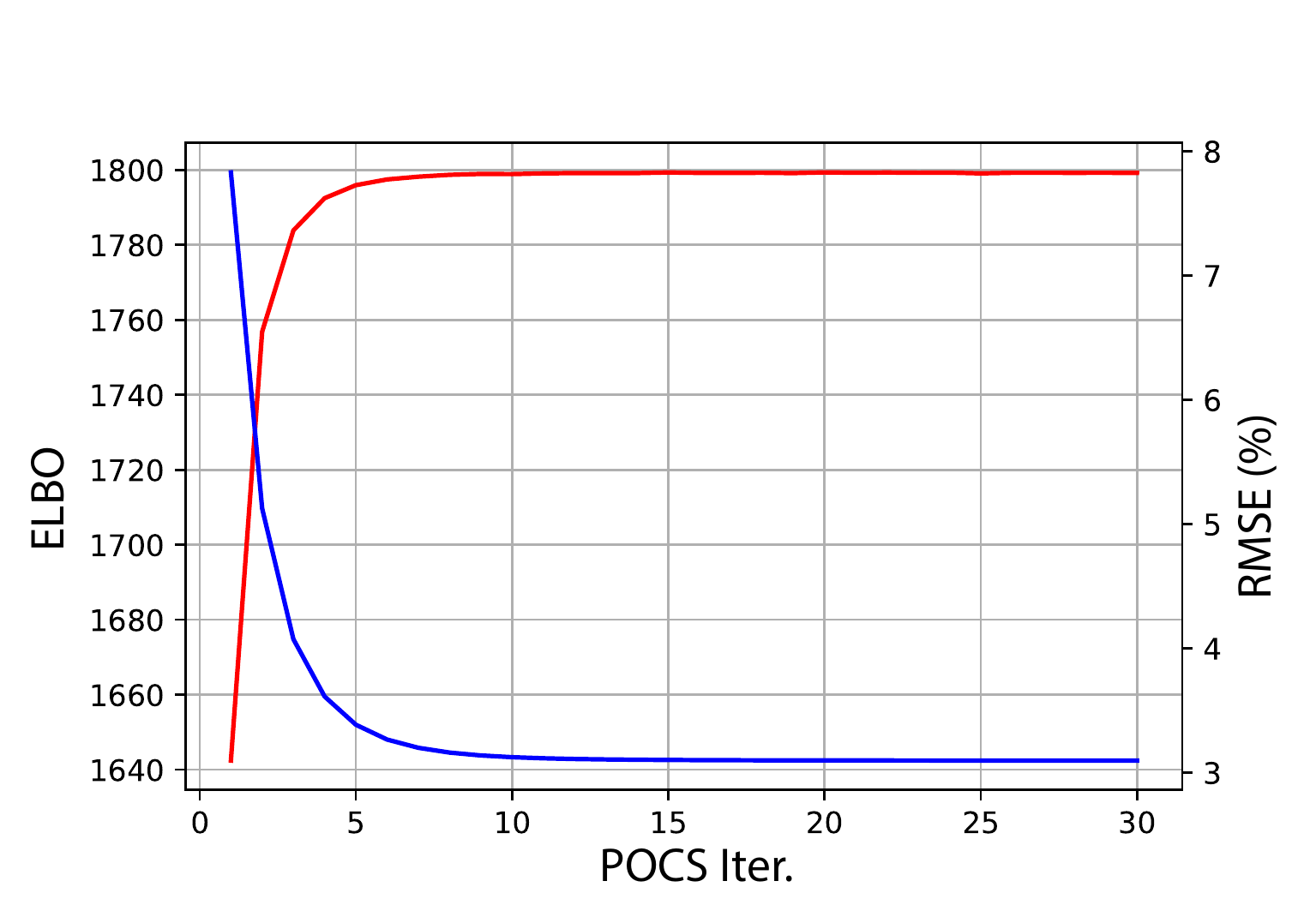}
\caption{ELBO (red) and RMSE (blue) values between an original and reconstructed HCP image during DDP reconstruction from an undersampled image with R=2. Convergence is achieved approximately after 15 iterations for R=2. For higher R values, more iterations are necessary. The RMSE starts with a high value as the iterations are initialized with the zero-filled image. Notice that ELBO also increases stably.}
\label{fig:convergence}
\end{figure}

Figure~\ref{fig:convergence} demonstrates the convergence of RMSE and ELBO values during iterations of the POCS algorithm for a random image from the test set for R=2. We observe the graceful decrease in RMSE and increase in ELBO, and convergence in both measures. 

\begin{figure}[ht]
\includegraphics[width=\textwidth]{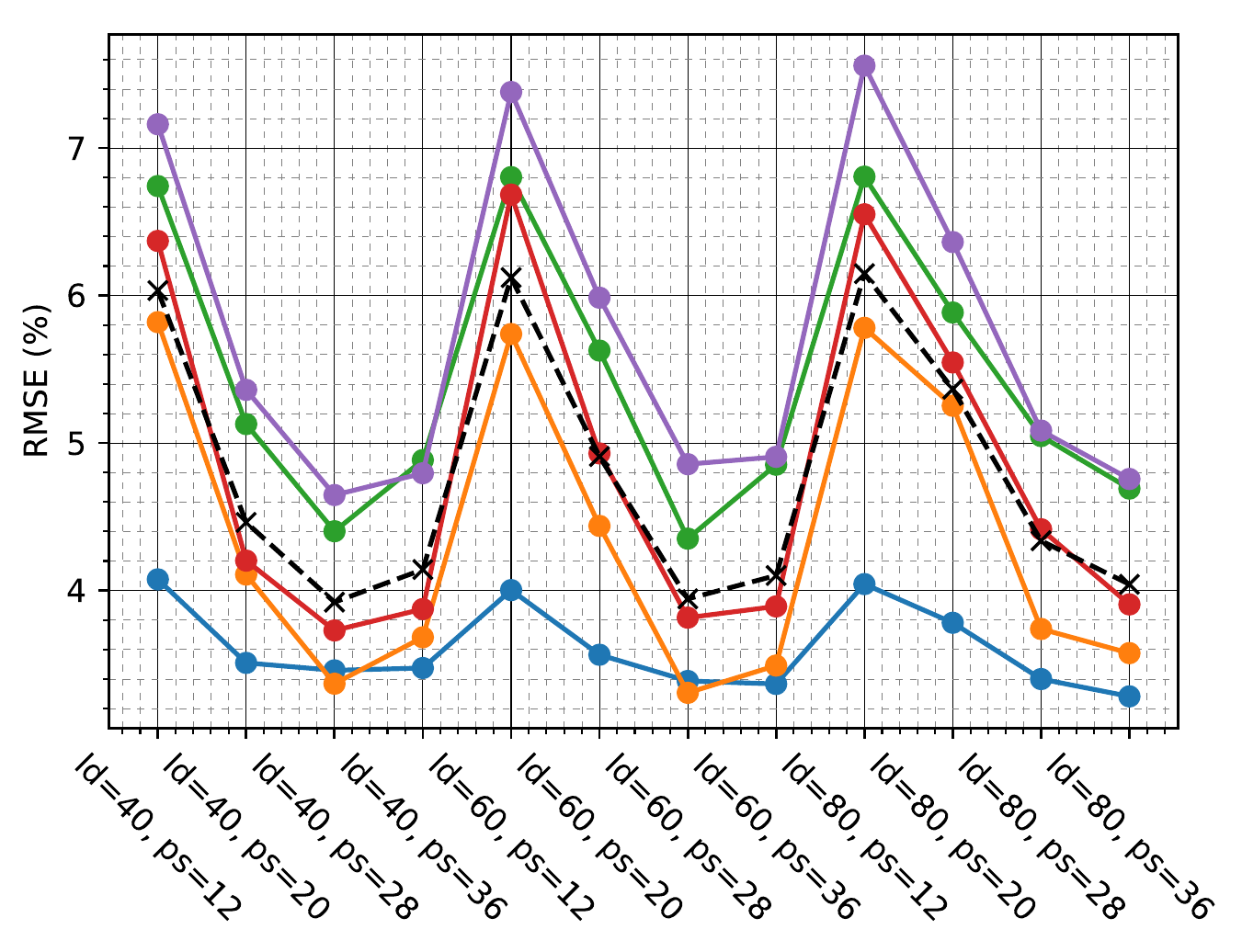}
\caption{RMSE values for varying the latent space dimension (ld) and patch size (ps) (5 subjects, R=3). The method performs similarly for different configurations indicating robustness to parameter selection to some extend. Each color indicates a different subject. The dashed line is the mean value for all subjects.}
\label{fig:ps_ld_rmse}
\end{figure}

\begin{figure}
\includegraphics[width=\textwidth]{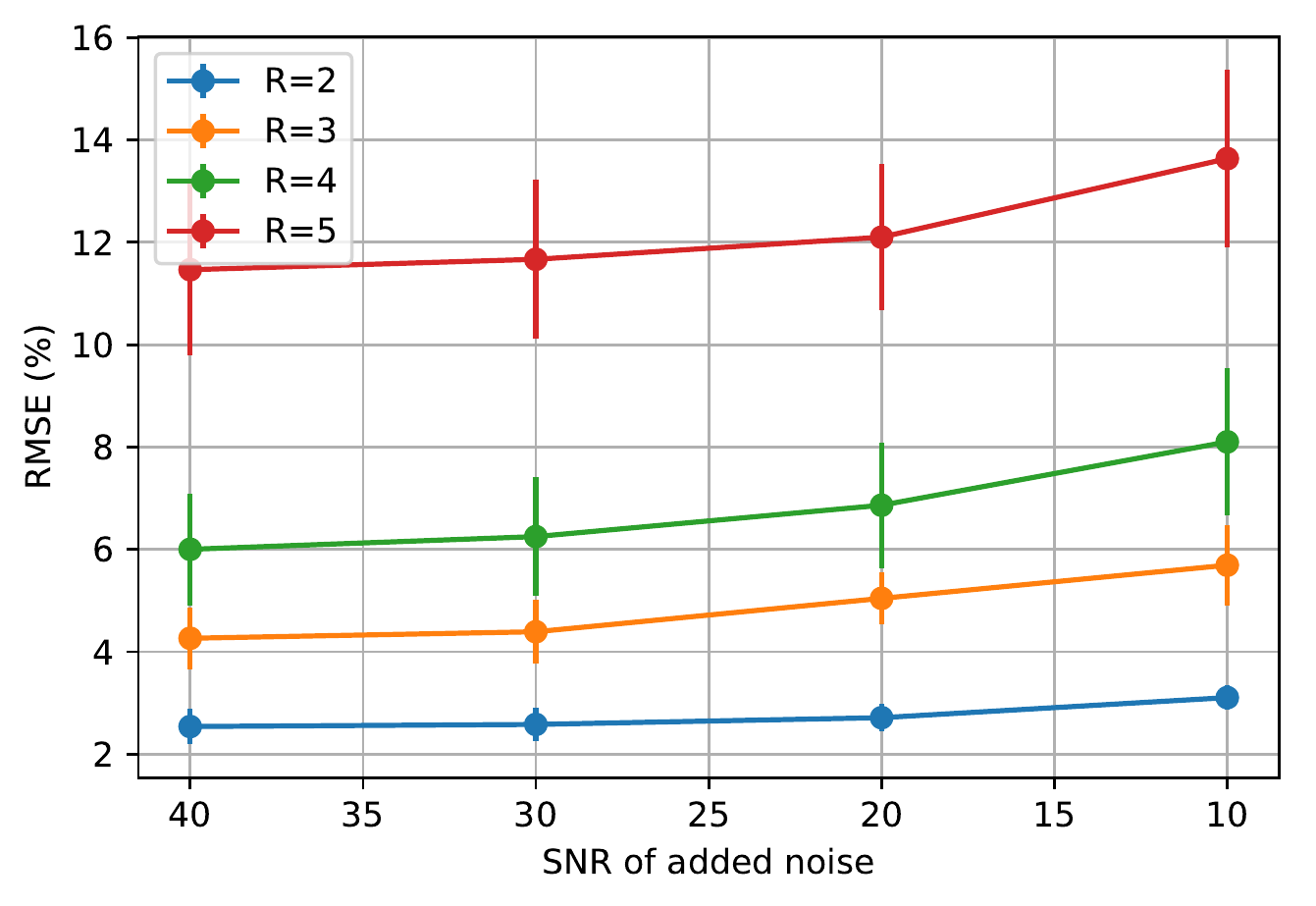}
\caption{RMSE vs SNR of added noise (5 subjects). As expected increasing noise increases the error in reconstruction. Notice the net SNR of the images is around half of the added value, due to the base noise in the original images. }
\label{fig:snr_vs_rmse}
\end{figure}

\begin{figure}[ht]
\includegraphics[width=\textwidth]{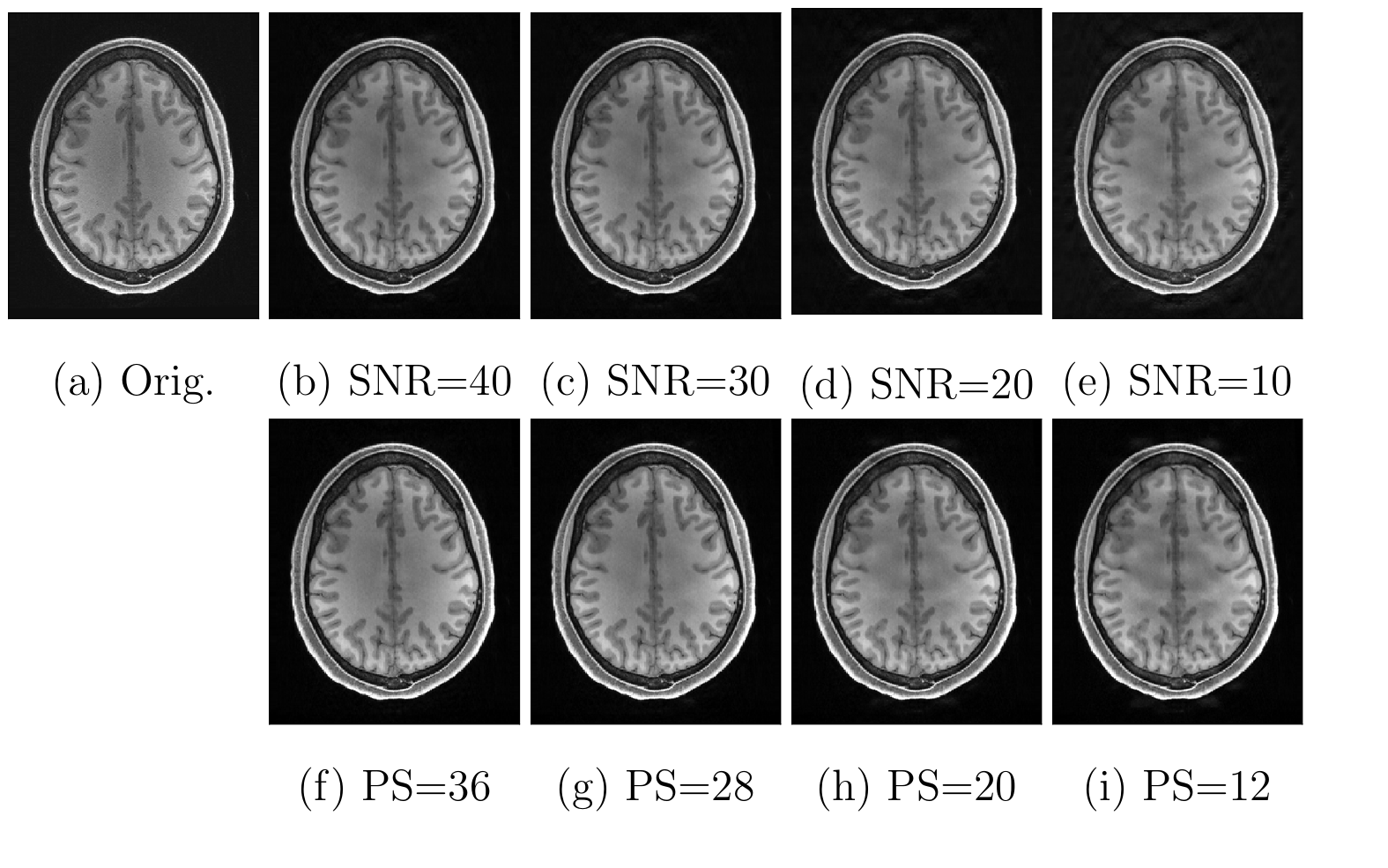}
\caption{Example reconstructions for varying SNR and patch size (PS).}
\label{fig:PS_snr_im}
\end{figure}

Figure~\ref{fig:ps_ld_rmse} shows results for different configurations of the model. We vary the dimensions of the latent space and the patch size to see how sensitive the method is to these design parameters. We do the experiments for 5 test subjects at R=3. We observe that patch-size has a much larger impact than latent-dimension. Reconstruction accuracy values do not change substantially with changing latent-dimension. On the other hand, we observe decrease in reconstruction error with decreasing patch-size, especially going from 12 to 20. This suggests the value of building explicit priors for larger image patches. Figure~\ref{fig:snr_vs_rmse} shows the decrease in performance with decreasing SNR in the undersampled image for 5 subjects for R=2 to 5. We calculate the RMSE between the noisy fully sampled image and the reconstruction. Increase in RMSE is observed, as expected, however, the rate of increase was slow for all the cases. Figure~\ref{fig:PS_snr_im} shows examples of reconstructions with varying patch size and SNR for R=3, and latent dimension 60.

\begin{figure}
\includegraphics[width=0.9\textwidth]{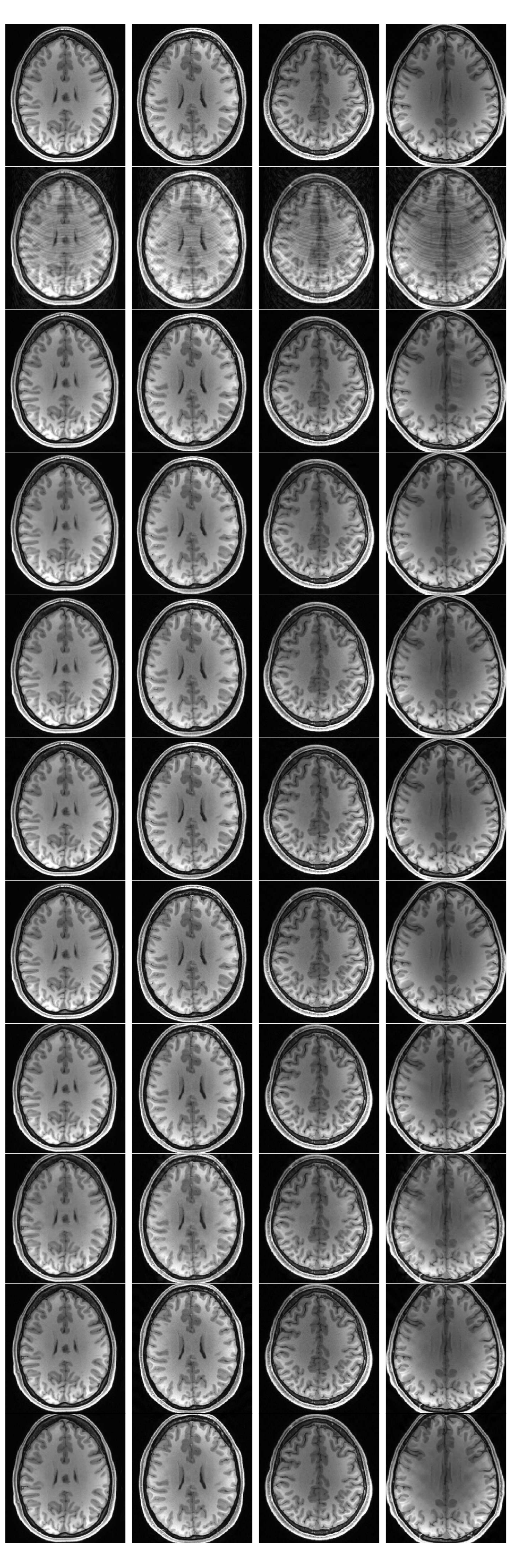}
\\[-3ex]
\caption{Reconstruction results from 4 different subjects with R=2. Rows show in top-down direction: the fully sampled images, zero-filling images, DDP, TV, DLMRI and ADMM-Net, BM3D-MR reconstructions for full FOV and reconstructions with the cropped FOV for DDP, SIDWT, FDLCP, PBDW.}
\label{fig:supp_rec_r2}
\end{figure}

\begin{figure}
\includegraphics[width=0.9\textwidth]{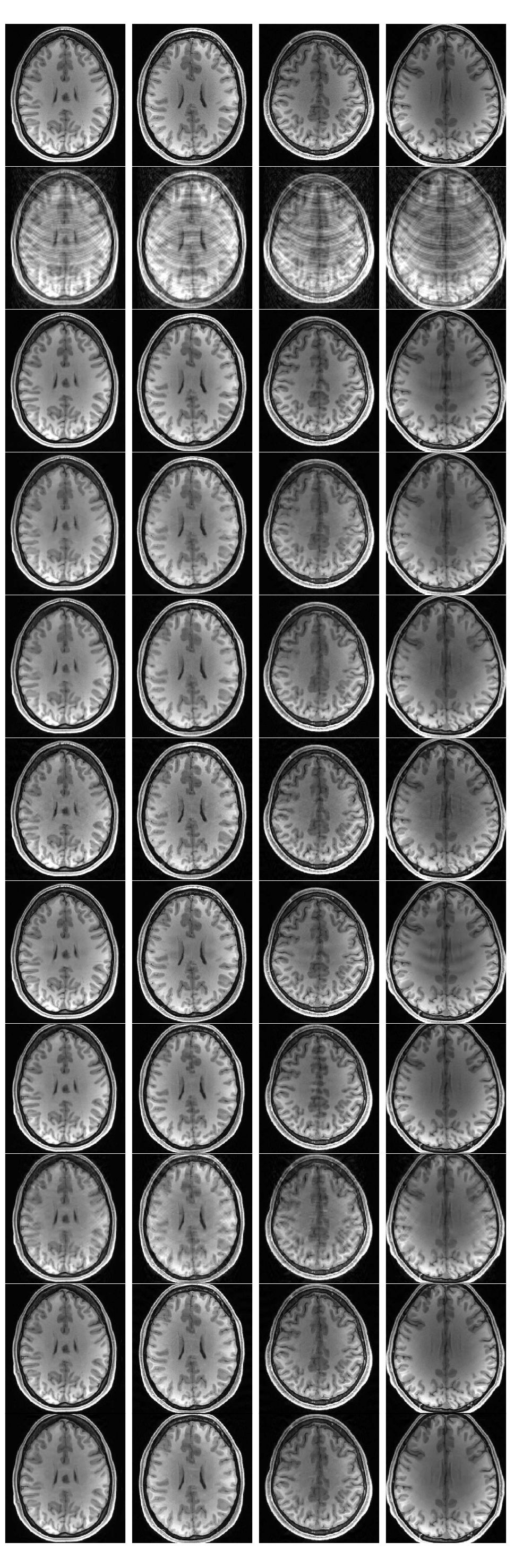}
\\[-3ex]
\caption{Similar display as in as Figure S7 with R=3}
\label{fig:supp_rec_r3}
\end{figure}

\begin{figure}
\includegraphics[width=1\textwidth]{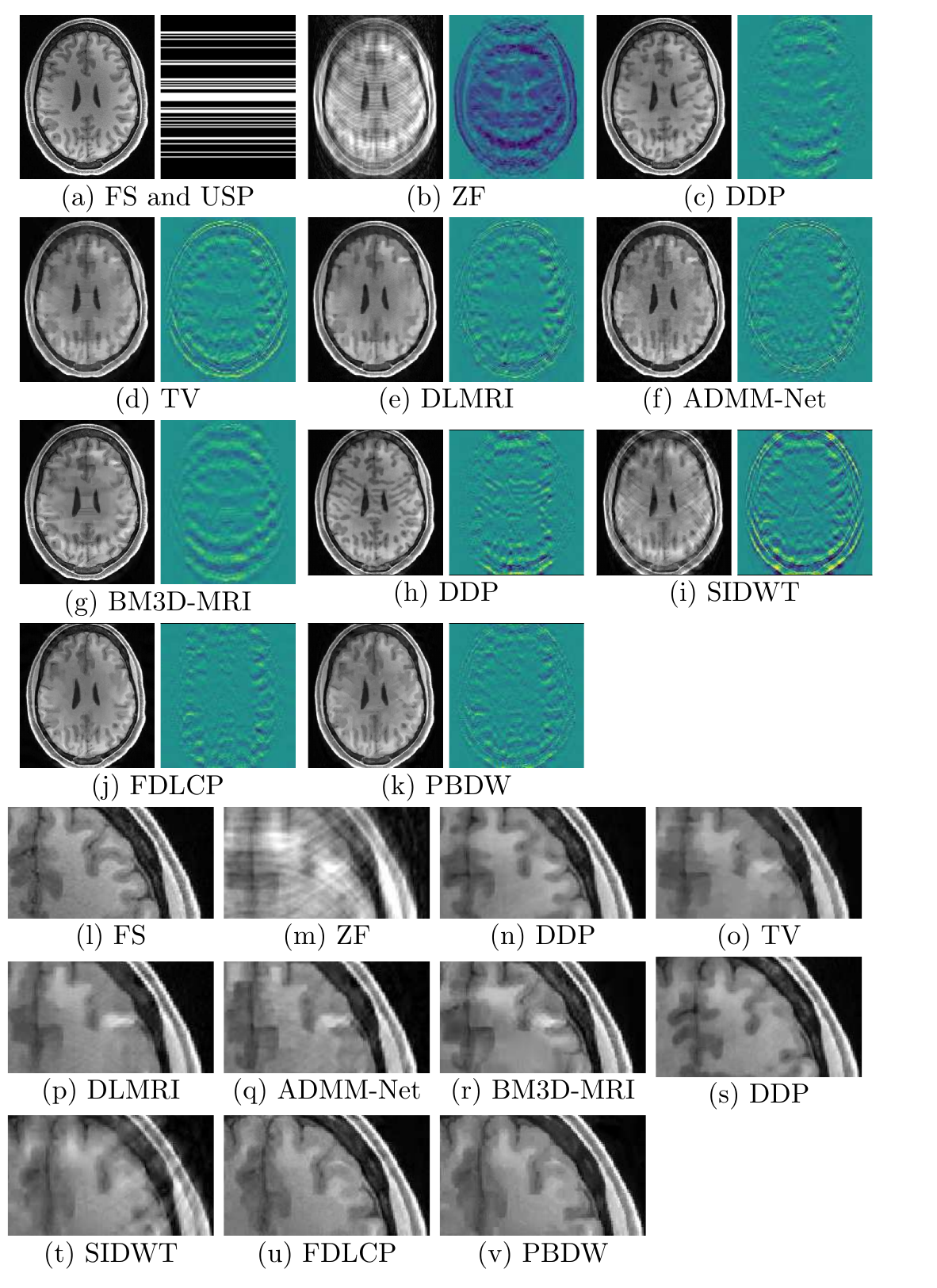}
\\[-3ex]
\caption{Similar display as in Figure 3  with R=5. }
\label{fig:recres_r5}
\end{figure}

Figures~\ref{fig:supp_rec_r2} and~\ref{fig:supp_rec_r3} present more images selected randomly from the test set and their reconstructions for R=2 and 3, respectively. Figures~\ref{fig:recres_r4} and~\ref{fig:recres_r5} presents the same image used in the main text and its reconstructions for R=4 and 5, respectively.

\begin{figure}
\includegraphics[width=1\textwidth]{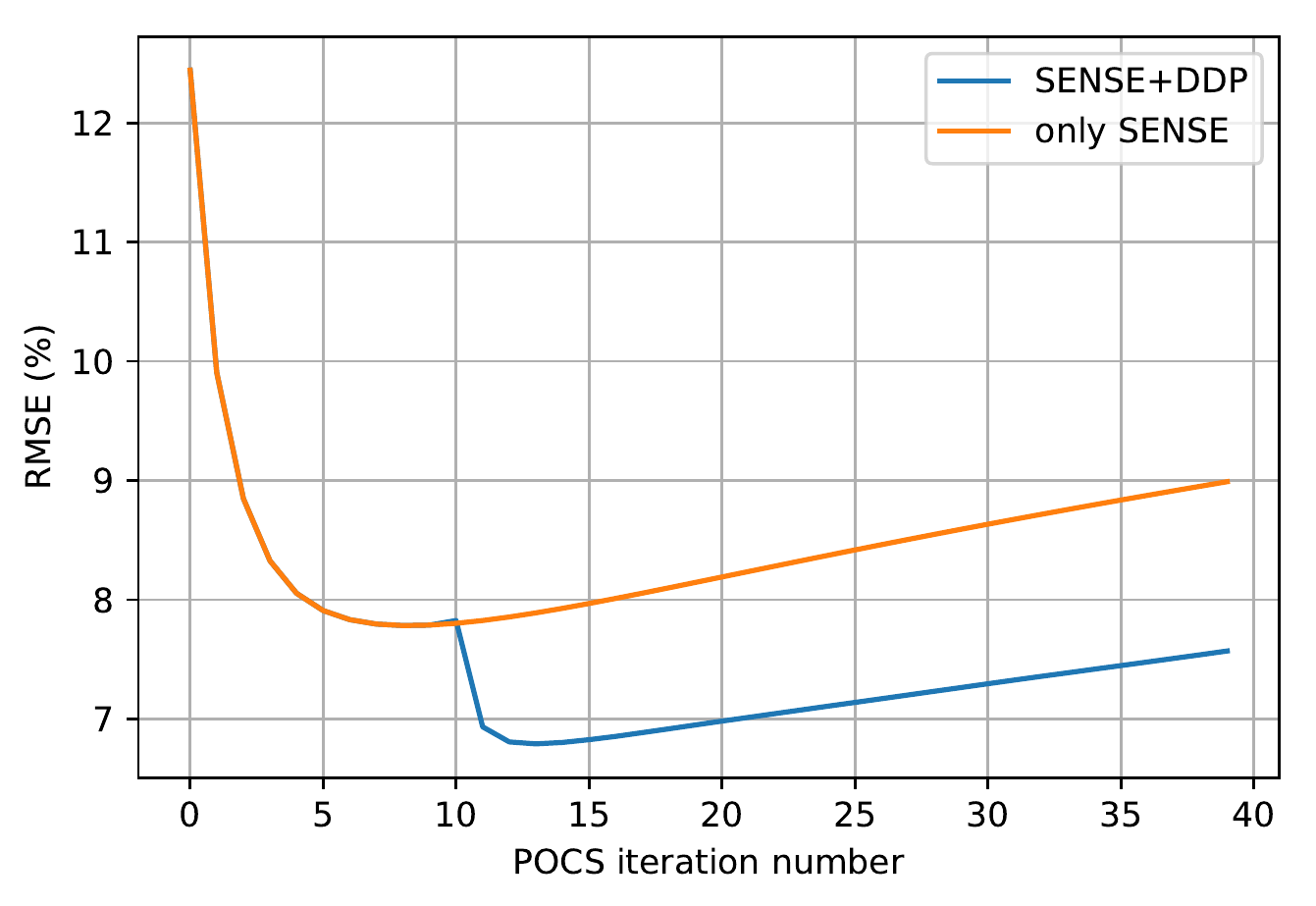}
\\[-3ex]
\caption{RMSE values for using doing SENSE reconstruction (i.e. only data consistency projections) vs doing combined SENSE and DDP reconstruction. Notice the DDP projection is switched on after the 10th iteration. The drop right after the first DDP projection demonstrates the added value due to the DDP projection. }
\label{fig:ddp_vs_sense_rmse}
\end{figure}

\begin{figure}
\includegraphics[width=1\textwidth]{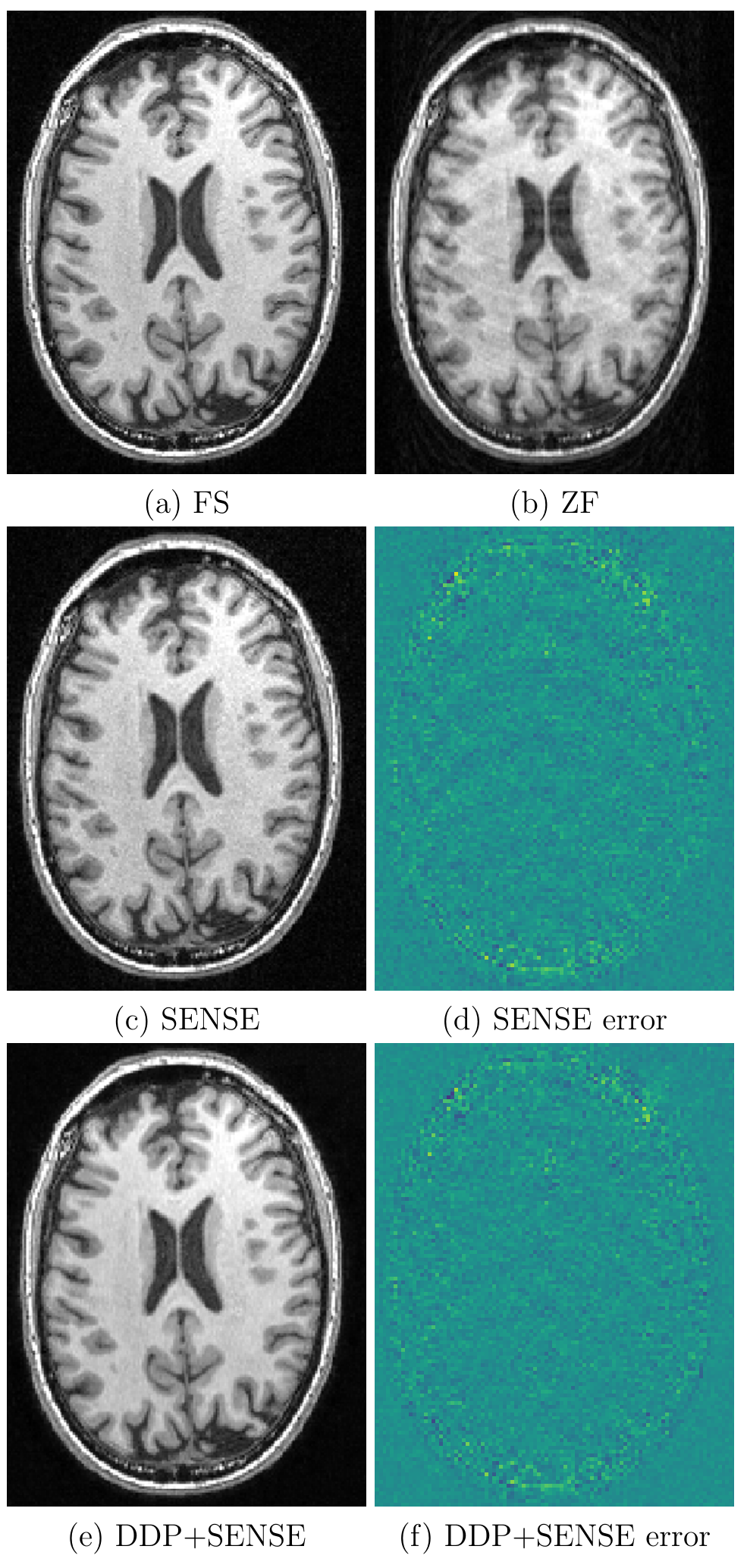}
\\[-3ex]
\caption{Reconstruction results for a measured subject with using only SENSE and the combination of SENSE and DDP. Error maps are clipped to (-0.3,0.3)}
\label{fig:ddp_vs_sense_ims}
\end{figure}

Figure~\ref{fig:ddp_vs_sense_rmse} shows the improvement achieved in RMSE by using DDP with SENSE in contrast to performing only SENSE reconstruction for one image acquired within this study. We implement the SENSE reconstruction by only doing data consistency steps as in~\cite{SamsonovPOCS}. Otherwise, we do the reconstruction as described in the main text, i.e. only data consistency projections for the first 10 iterations, then switch on the DDP projection. We show only 40 iterations. The drop in the RMSE value after iteration 10, i.e. right after the DDP projection is applied, shows the added value of doing the DDP projection. We also observe increase in RMSE for both reconstruction methods with increasing iterations. This increase is attributed to the discrepancy between the true and used coil sensitivities. Reconstructed images for this subject are given in Figure~\ref{fig:ddp_vs_sense_ims}. We observe larger aliasing artifacts in only SENSE reconstruction. 

\begin{figure*}[!htb]
\includegraphics[trim={0 0 0 0},clip,width=0.9\textwidth]{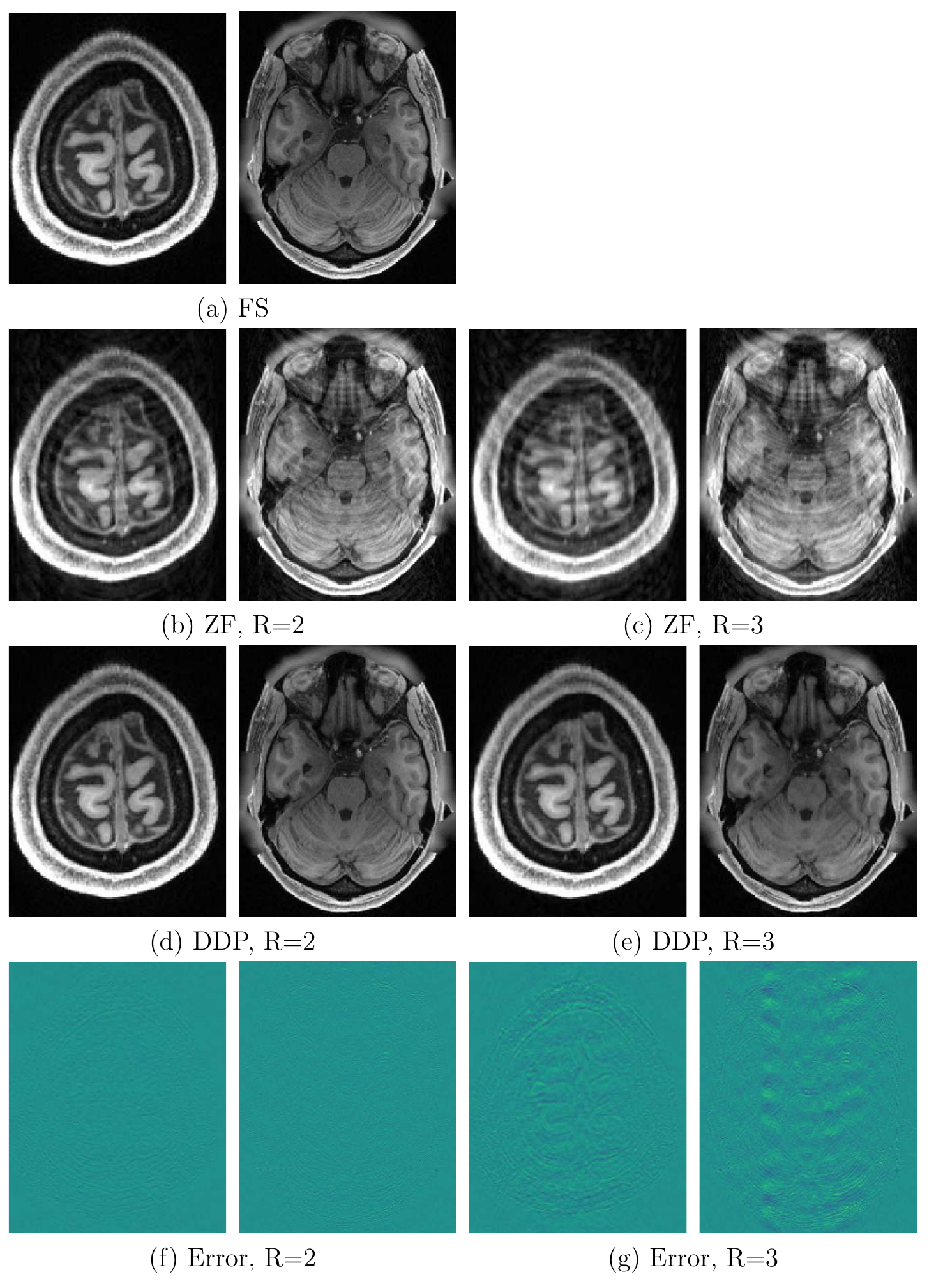}
\caption{DDP reconstruction results (R=2 and 3) of two slices from anatomical areas, on which the prior was not trained. The left and right columns show slices from superior and inferior regions, respectively. Error maps are clipped to (-0.3,0.3). }
\label{fig:offc_r2}
\end{figure*}

In Figure~\ref{fig:offc_r2} we show results for two slices from the inferior and superior regions of the brain, on which the prior model was not trained. Visual inspection shows that the superior slices are reconstructed fairly well, whereas the model struggles more in the inferior slices, especially in the cerebellum. The RMSE values for both anatomical areas together are 4.28\% (0.68) and 6.90\% (1.23) (N=17) for R=2 and 3, respectively. These values show an overall decrease in performance, as expected, when the prior is used for reconstructing slices from regions which are not represented in the training set. This decrease might also be partly due to the fact that the inferior slices generally contain more complex structures than central slices (e.g. the cerebellum). Furthermore, the RMSE values are higher for the inferior slices, supporting the visual observations: the mean (std) RMSE values for superior slices are  3.66\% (0.44) and 5.43\% (0.54) (N=6) in contrast to 4.44\% (0.71) and 7.41\% (1.06) (N=11) for inferior slices for R=2 and 3, respectively. This difference is highly likely due to the difference in the structural complexity between the regions. 

\begin{figure*}[!htb]
\includegraphics[trim={0 0 0 0},clip,width=0.99\textwidth]{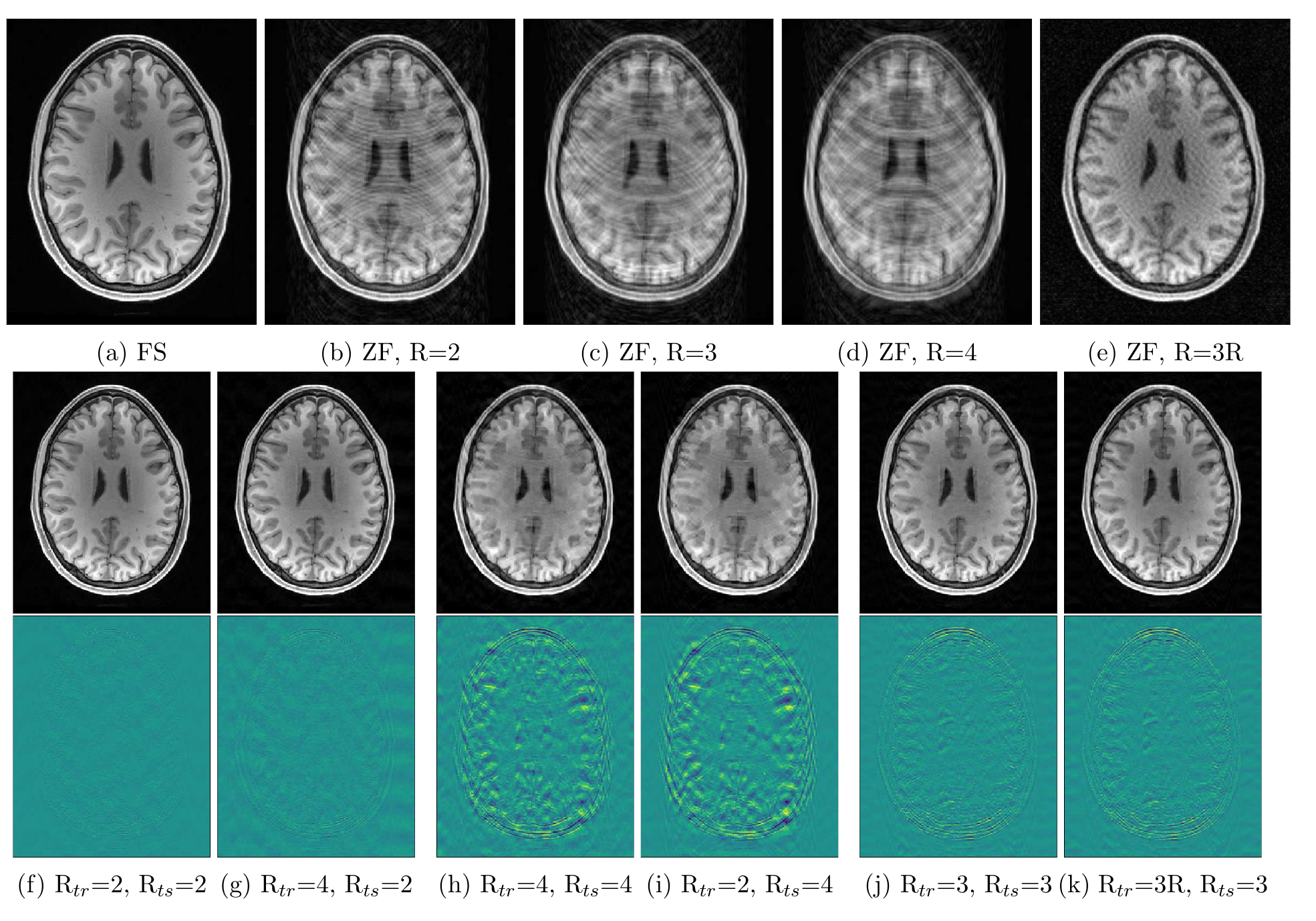}
\caption{ADMM-Net reconstruction results with differing training and test undersampling patterns. The first row shows the fully sampled image and zero-filled images. R denotes the Cartesian undersampling ratio used for generating the zero-filled images. The second and third rows show the reconstruction results and the error maps, respectively. R$_{tr}$ and R$_{ts}$ stand for Cartesian undersampling factors for training and testing, respectively. "3R" denotes pseudo-radial undersampling pattern with factor 3. Error maps are clipped to (-0.3,0.3). }
\label{fig:admmnet_images}
\end{figure*}

In Figure~\ref{fig:admmnet_images} we show ADMM-Net reconstruction results for a slice in the setting where training and test undersampling schemes differ. The image pairs (f-g), (h-i) and (j-k) correspond to visual results for experiments I, II and III in Table II in the main text, respectively. We also show the zero-filled images to give an idea on the difficulty of the reconstruction problem.


\newpage

\begin{figure}
\includegraphics[width=1\textwidth]{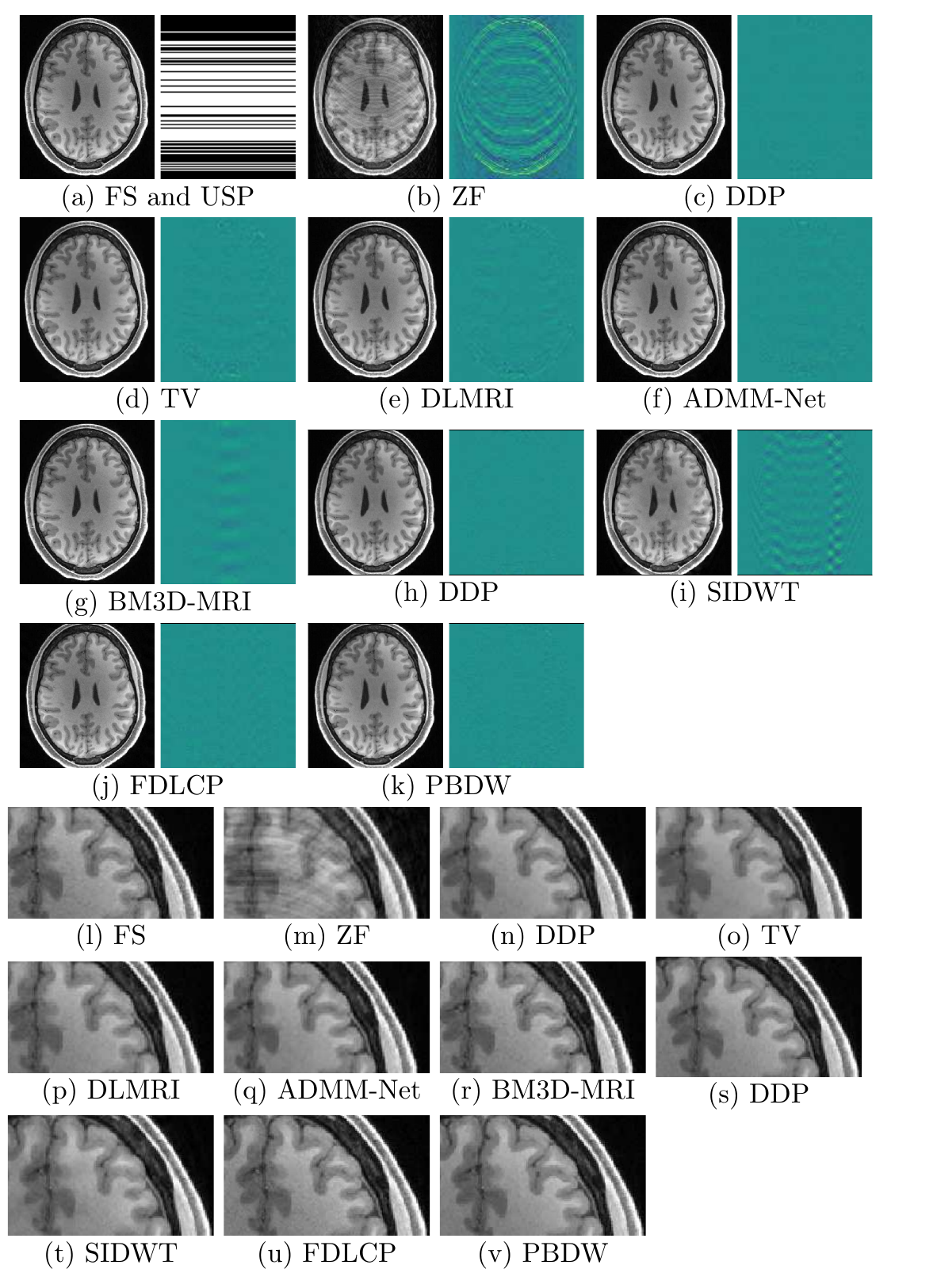}
\\[-3ex]
\caption{Similar display as in Figure 3  with R=2. }
\label{fig:recres_r4}
\end{figure}

\begin{figure}
\includegraphics[width=1\textwidth]{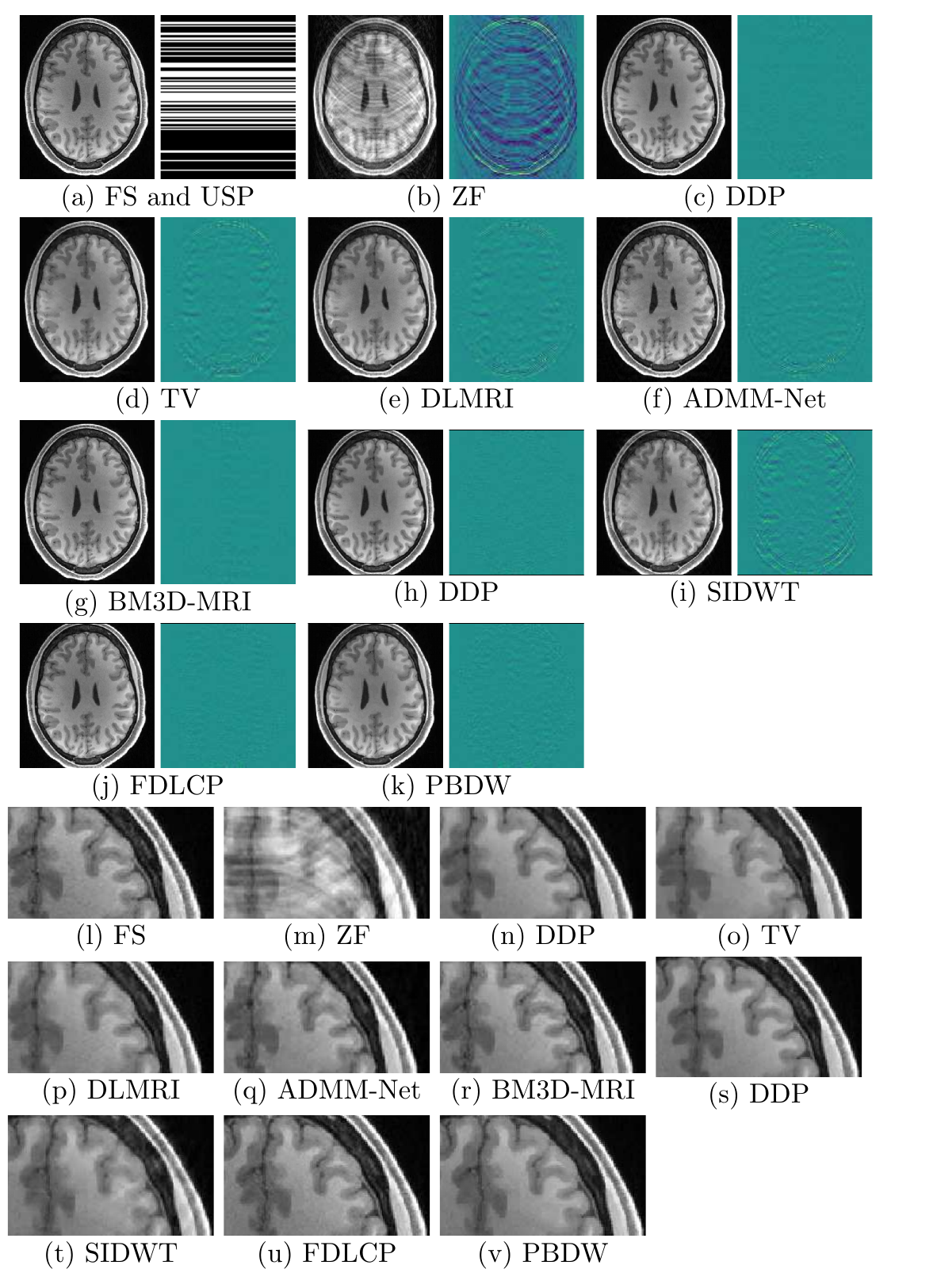}
\\[-3ex]
\caption{Similar display as in Figure 3  with R=4. }
\label{fig:recres_r4}
\end{figure}

\newpage


\end{document}